\documentclass[twoside,twocolumn, 10pt]{article}
\usepackage{microtype}
\usepackage[margin=0.52in,top=30mm,bottom=40mm,columnsep=15pt]{geometry}
\usepackage[english]{babel} % Language hyphenation and typographical rules
\RequirePackage{fix-cm}
\usepackage{authblk}
\usepackage{mathptmx}      % use Times fonts if available on your TeX system
\usepackage{enumitem}
\usepackage{array,multirow,graphicx}
\DeclareMathAlphabet{\pazocal}{OMS}{zplm}{m}{n}
\usepackage{amsmath,amssymb}
\usepackage{textcomp}
\usepackage{subcaption}
\usepackage{gensymb}
\usepackage{url}
\usepackage{caption}

\date{}

\usepackage{wrapfig}

\usepackage[english]{babel}
\newcommand\figref{Fig.~\ref}
\newcommand\tabref{Table~\ref}
\newcommand\eref{Eq. ~\ref}
\newcommand\bt{\textbf}
\newcommand\eg{\emph{e.g. }}

\newcommand\etal{\emph{et al. }}
\DeclareMathOperator{\EX}{\mathbb{E}}% expected value
\begin{document}
\title{Realistic Speech-Driven Facial Animation with GANs}

\author[1]{Konstantinos Vougioukas}
\author[1,2]{Stavros Petridis}
\author[1,2]{Maja Pantic}
\affil[1]{iBUG Group, Imperial College London}
\affil[2]{Samsung AI Centre, Cambridge, UK}

\maketitle
\begin{abstract}
\noindent Speech-driven facial animation is the process that automatically synthesizes talking characters based on speech signals. The majority of work in this domain creates a mapping from audio features to visual features. This approach often requires post-processing using computer graphics techniques to produce realistic albeit subject dependent results. We present an end-to-end system that generates videos of a talking head, using only a still image of a person and an audio clip containing speech, without relying on handcrafted intermediate features. Our method generates videos which have (a) lip movements that are in sync with the audio and (b) natural facial expressions such as blinks and eyebrow movements. Our temporal GAN uses 3 discriminators focused on achieving detailed frames, audio-visual synchronization and realistic expressions. We quantify the contribution of each component in our model using an ablation study and we provide insights into the latent representation of the model. The generated videos are evaluated based on sharpness, reconstruction quality, lip-reading accuracy, synchronization as well as their ability to generate natural blinks.
\end{abstract}

\section{Introduction}
\label{intro}
Computer Generated Imagery (CGI) has become an inextricable part of the entertainment industry due to its ability to produce high quality results in a controllable manner. One very important element of CGI is facial animation because the face is capable of conveying a plethora of information not only about the character but also about the scene in general (\eg tension, danger). The problem of generating realistic talking heads is multifaceted, requiring high-quality faces, lip movements synchronized with the audio, and plausible facial expressions. This is especially challenging because humans are adept at picking up subtle abnormalities in facial motion and audio-visual synchronization.

Facial synthesis in CGI is traditionally performed using face capture methods, which have seen drastic improvement over the past years and can produce faces that exhibit a high level of realism. However, these approaches require expensive equipment and significant amounts of labour, which is why CGI projects are still mostly undertaken by large studios. In order to drive down the cost and time required to produce high quality CGI researchers are looking into automatic face synthesis using machine learning techniques. Of particular interest is speech-driven facial animation since speech acoustics are highly correlated with facial movements \cite{Yehia1998QuantitativeBehavior}.

These systems could simplify the film animation process through automatic generation from the voice acting. They can also be applied in post-production to achieve better lip-synchronization in movie dubbing. Moreover, they can be used to generate parts of the face that are occluded or missing in a scene. Finally, this technology can improve band-limited visual telecommunications by either generating the entire visual content based on the audio or filling in dropped frames.

The majority of research in this domain has focused on mapping audio features (\eg MFCCs) to visual features (\eg landmarks, visemes) and using computer graphics (CG) methods to generate realistic faces \cite{Karras2017}. Some methods avoid the use of CG by selecting frames from a person-specific database and combining them to form a video \cite{Bregler1997, Suwajanakorn2017}. Regardless of which approach is adopted these methods are subject dependent and are often associated with a considerable overhead when transferring to new speakers.

Subject independent approaches have been proposed that transform audio features to video frames \cite{Chung2017,  Chen2018LipGlance}. However, most of these methods restrict the problem to generating only the mouth. Even techniques that generate the entire face are primarily focused on obtaining realistic lip movements, and typically neglect the importance of generating facial expressions. Natural facial expressions play a crucial role in producing truly realistic characters and their absence creates an unsettling feeling for many viewers. This lack of expressions is a clear tell-tale sign of generated videos which is often exploited by systems such as the one proposed in \cite{LiInBlinking}, which exposes synthetic faces based on the existence and frequency of blinks.

Some methods generate frames based solely on present information \cite{Chung2017, Chen2018LipGlance}, without taking into account the dynamics of facial motion. However, generating natural sequences, which are characterized by a seamless transition between frames, can be challenging when using this static approach. Some video generation methods have dealt with this problem by generating the entire sequence at once \cite{Vondrick2016} or in small batches \cite{Saito2016}. However, this introduces a lag in the generation process, prohibiting their use in real-time applications and requiring fixed length sequences for training.

\begin{figure}
    \includegraphics[width=\columnwidth]{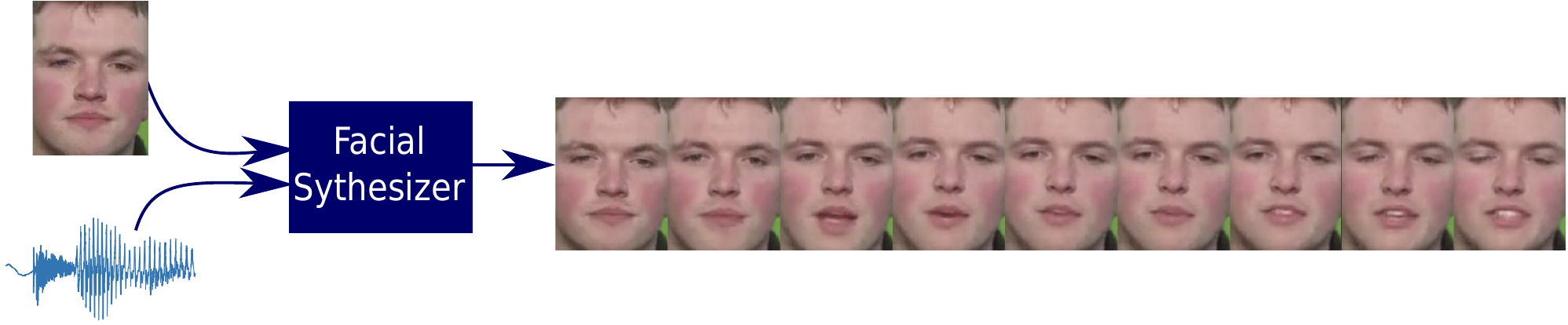}
    \caption{The proposed end-to-end face synthesis model, capable of producing realistic sequences of faces using one still image and an audio track containing speech. The generated sequences exhibit smoothness and natural expressions such as blinks and frowns.}
    \label{fig:1}
\end{figure}

In this work we propose a temporal generative adversarial network (GAN)\footnote{Videos and models are available on the following website: \\ \url{https://sites.google.com/view/facial-animation}}, capable of generating a video of a talking head from an audio signal and a single still image (see \figref{fig:1}). Our model builds on the system proposed in \cite{Vougioukas2018End-to-EndGANs} which uses separate discriminators at the frame and sequence levels to generate realistic videos. The frame-level discriminator ensures that generated frames are sharp and detailed, whereas the temporal discriminator is responsible for audio visual correspondence and generating realistic facial movements. During training the discriminator learns to differentiate real and fake videos based on synchrony or the presence of natural facial expressions. Although the temporal discriminator helps with the generation of expressions and provides a small improvement in audio-visual correspondence, there is no way of ensuring that both these aspects are captured in the video.

To solve this problem we propose using 2 temporal discriminators to enforce audio-visual correspondence and realistic facial movements on the generated videos. By separating these two tasks, which were undertaken by a single discriminator in \cite{Vougioukas2018End-to-EndGANs}, we are able to explicitly focus on audio-visual synchronization through a synchronisation discriminator trained to detect audio-visual misalignment. Furthermore, isolating expressions from synchronisation further encourages the generation of spontaneous facial expressions, such as blinks.

We also present a comprehensive assessment of the performance of our method. This is done using a plethora of quantitative measures and an in depth analysis that is missing from previous studies. Our model is trained and evaluated on the GRID \cite{Cooke2006}, TCD TIMIT \cite{Harte2015}, CREMA-D \cite{Cao2014CREMA-D:Dataset} and LRW \cite{Chung16} datasets. 

The frame quality is measured using well-established reconstruction and sharpness metrics. Additionally, we use lip reading systems to verify the accuracy of the spoken words and face verification to ensure that the identity is correctly captured and maintained throughout the sequence. Furthermore, we examine the audio-visual correspondence in produced videos by using a recent speech synchronization detection method. Finally, using a blink detector we measure the number of blinks on the generated videos as well as the blink duration.

This work provides an in-depth look at our method, examining how each element affects the quality of the video. The contribution of each discriminator in our GAN is quantified using the aforementioned metrics through an ablation study performed on the GRID \cite{Cooke2006} dataset. Furthermore, we examine the latent space in order to determine how well our system encodes the speaker identity. Moreover, we analyze the characteristics of the spontaneous expressions on videos generated using our method and compare with those of real videos. Finally, we present the results of an online Turing test, where users are shown a series of generated and real videos and are asked to identify the real ones.

\section{Related Work}
\label{sec:related}
The problem of speech-driven video synthesis is not new in computer vision and in fact, has been a subject of interest for decades. Yehia \etal \cite{Yehia1998QuantitativeBehavior} were first to investigate the relationship between acoustics, vocal-tract and facial motion, discovering a strong correlation between visual and audio features and a weak coupling between head motion and the fundamental frequency of the speech signal \cite{Yehia2002LinkingAcoustics}. These findings have encouraged researchers to find new ways to model the audio-visual relationship. The following sections present the most common methods used in each modelling approach.

\subsection{Visual Feature Selection and Blending}
\label{sec:vis_feat_selection}
The relationship between speech and facial motion has been exploited by some CG methods, which assume a direct correspondence between basic speech and video units. Cao \etal \cite{Cao2005ExpressiveAnimation} build a graph of visual representations called \textit{animes} which correspond to audio features. The graph is searched in order to find a sequence that best represents a given utterance under certain co-articulation and smoothness constraints. Additionally, this system learns to detect the emotion of the speech and adjust the \textit{animes} accordingly to produce movements on the entire face. The final result is obtained by time-warping the \textit{anime} sequence to match the timing of the spoken utterance and blending for smoothness. Such methods use a small set of visual features and interpolate between key frames to achieve smooth movement. This simplification of the facial dynamics usually results in unnatural lip movements, which is why methods that attempt to model the facial dynamics are preferred over these approaches.

\subsection{Synthesis Based on Hidden Markov Models}
\label{sec:hmm}
Some of the earliest methods for facial animation relied on Hidden Markov Models (HMMs) to capture the dynamics of the video and speech sequences. Simons and Cox \cite{Simons1990} used vector quantization to achieve a compact representation of video and audio features, which were used as the states for their fully connected Markov model. The Viterbi algorithm was used to recover the most likely sequence of mouth shapes for a speech signal. A similar approach is used in \cite{Yamamoto1998} to estimate the sequence of lip parameters. Finally, the {\em Video Rewrite} method \cite{Bregler1997} relies on the same principles to obtain a sequence of triphones, which are used to look up mouth images from a database. The final result is obtained by time-aligning the images to the speech and then spatially aligning and stitching the jaw sections to the background face.

Since phonemes and visemes do not have a one-to-one correspondence some HMM-based approaches replace the single Markov chain approach with a multi-stream approach. Xie \etal \cite{Xie2007AAnimation} propose a coupled HMM to model the audio-visual dependencies and compare the performance of this model to other single and multi-stream HMM architectures.

\subsection{Synthesis Based on Deep Neural Networks}
\label{sec:deep_neural_networks}
Although HMMs were initially preferred to neural networks due to their explicit breakdown of speech into intuitive states, recent advances in deep learning have resulted in neural networks being used in most modern approaches. Like past attempts, most of these methods aim at performing a feature-to-feature translation. A typical example of this, proposed in \cite{Taylor2017}, uses a deep neural network (DNN) to transform a phoneme sequence into a sequence of shapes for the lower half of the face. Using phonemes instead of raw audio ensures that the method is subject independent. 

Most deep learning approaches use convolutional neural networks (CNN) due to their ability to efficiently capture useful features in images. Karras \etal \cite{Karras2017} use CNNs to transform audio features to 3D meshes of a specific person. This system is conceptually broken into sub-networks responsible for capturing articulation dynamics and estimating the 3D points of the mesh. 

Analogous approaches,which are capable of generating facial descriptors from speech using recurrent neural networks (RNNs) have been proposed in \cite{Fan2015, Suwajanakorn2017, Pham2017Speech}. In particular, the system proposed in \cite{Suwajanakorn2017} uses Long Short Term Memory (LSTM) cells to produce mouth shapes from Mel-Frequency Cepstral Coefficients (MFCCs). For each generated mouth shape a set of best matching frames is found from a database and used to produce mouth images. These mouth shapes are blended with the frames of a real target video to produce very realistic results. 

Although visual features such as mouth shapes and 3D meshes are very useful for producing high quality videos they are speaker specific. Therefore, methods that rely on them are subject dependent and require additional retraining or re-targeting steps to adapt to new faces. For this reason methods like the one proposed in \cite{Zhou2018VisemeNet:Animation} use speaker independent features such as visemes and Jaw and Lip (JALI) parameters.

Finally, Chung {\em et al.} \cite{Chung2017} proposed a CNN applied on MFCCs that generates subject independent videos from an audio clip and a still frame. The method uses an $L_1$ loss at the pixel level resulting in blurry frames, which is why a deblurring step is also required. Secondly, this loss at the pixel level penalizes any deviation from the target video during training, providing no incentive for the model to produce spontaneous expressions and resulting in faces that are mostly static except for the mouth. 

\subsection{GAN-Based Video Synthesis}
\label{sec:gans}
The recent introduction of GANs in \cite{Goodfellow2014} has shifted the focus of the machine learning community to generative modelling. GANs consist of two competing networks: a generative network and a discriminative network. The generator's goal is to produce realistic samples and the discriminator's goal is to distinguish between the real and generated samples. This competition eventually drives the generator to produce highly realistic samples. GANs are typically associated with image generation since the adversarial loss produces sharper, more detailed images compared to $L_1$ and  $L_2$ losses. However, GANs are not limited to these applications and can be extended to handle videos \cite{Mathieu2015, Li2017, Vondrick2016, Tulyakov2017}. 

Straight-forward adaptations of GANs for videos are proposed in \cite{Vondrick2016, Saito2016}, replacing the 2D convolutional layers with 3D convolutional layers. Using 3D convolutions in the generator and discriminator networks is able to capture temporal dependencies but requires fixed length videos. This limitation was overcome in \cite{Saito2016} but constraints need to be imposed in the latent space to generate consistent videos. CNN based GAN approaches have been used for speech to video approaches such as the one proposed in \cite{zhou2019talking}.

The {\em MoCoGAN} system proposed in \cite{Tulyakov2017} uses an RNN-based generator, with separate latent spaces for motion and content. This relies on the empirical evidence shown in \cite{Radford2015} that GANs perform better when the latent space is disentangled. {\em MoCoGAN} uses a 2D and 3D CNN discriminator to judge frames and sequences respectively. A sliding window approach is used so that the 3D CNN discriminator can handle variable length sequences. Furthermore, the GAN-based system proposed in \cite{Pham2018GenerativeNetwork} uses Action Unit (AU) coefficients to animate a head. A similar approach is used in the GANimation model proposed in \cite{pumarola2018ganimation}. These approaches can be combined with speech-driven animation methods \cite{Pham2017Speech} that produce AU coefficients which drive facial expressions from speech.

GANs have also been used in a variety of cross-modal applications, including text-to-video and audio-to-video. The text-to-video model proposed in \cite{Li2017} uses a combination of variational auto encoders (VAE) and GANs in its generating network and a 3D CNN as a sequence discriminator. Finally, Chen {\em et al.} \cite{Chen2017} propose a GAN-based encoder-decoder architecture that uses CNNs in order to convert audio spectrograms to frames and vice versa. This work is extended in \cite{Chen_2019_CVPR}, using an attention mechanism which helps the network focus on frame regions that correlate highly with the audio. However as a result this method neglects other areas such as the brow and eyes.

\section{Speech-Driven Facial Synthesis}
\label{sec:method}
\begin{figure}[t]
    \centering
    \includegraphics[width=0.9\columnwidth]{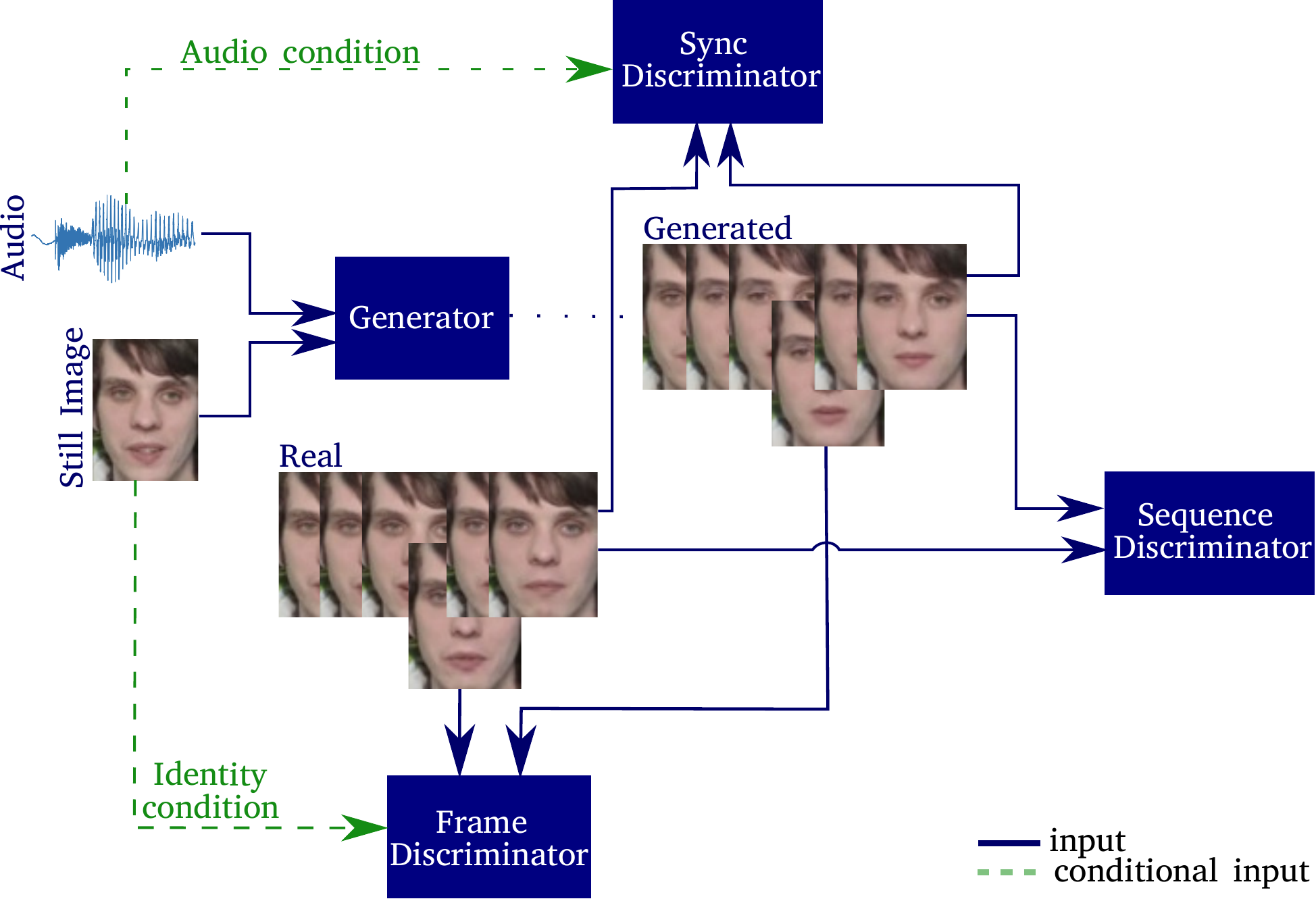}
    \caption{The deep model for speech-driven facial synthesis. It uses 3 discriminators to incorporate different aspects of a realistic video.}
    \label{fig:model_block}
\end{figure}

The proposed architecture for speech-driven facial synthesis is shown in \figref{fig:model_block}. The system consists of a temporal generator and multiple discriminators, each of which evaluates the generated sequence from a different perspective. The capability of the generator to capture various aspects of natural sequences is proportional to the ability of each discriminator to discern videos based on them.

\subsection{Generator}
\label{sec:generator}
The generator accepts as input a single image and an audio signal, which is divided into overlapping frames corresponding to $0.2$ seconds. Each audio frame must be centered around a video frame. In order to achieve this one-to-one correspondence we zero pad the audio signal on both sides and use the following formula for the stride:

\begin{equation}
stride=\frac{audio\; sampling\;rate}{video\;fps}
\label{eq:cutting_stride}
\end{equation}

\begin{figure}[t]
    \centering
    \includegraphics[width=\columnwidth]{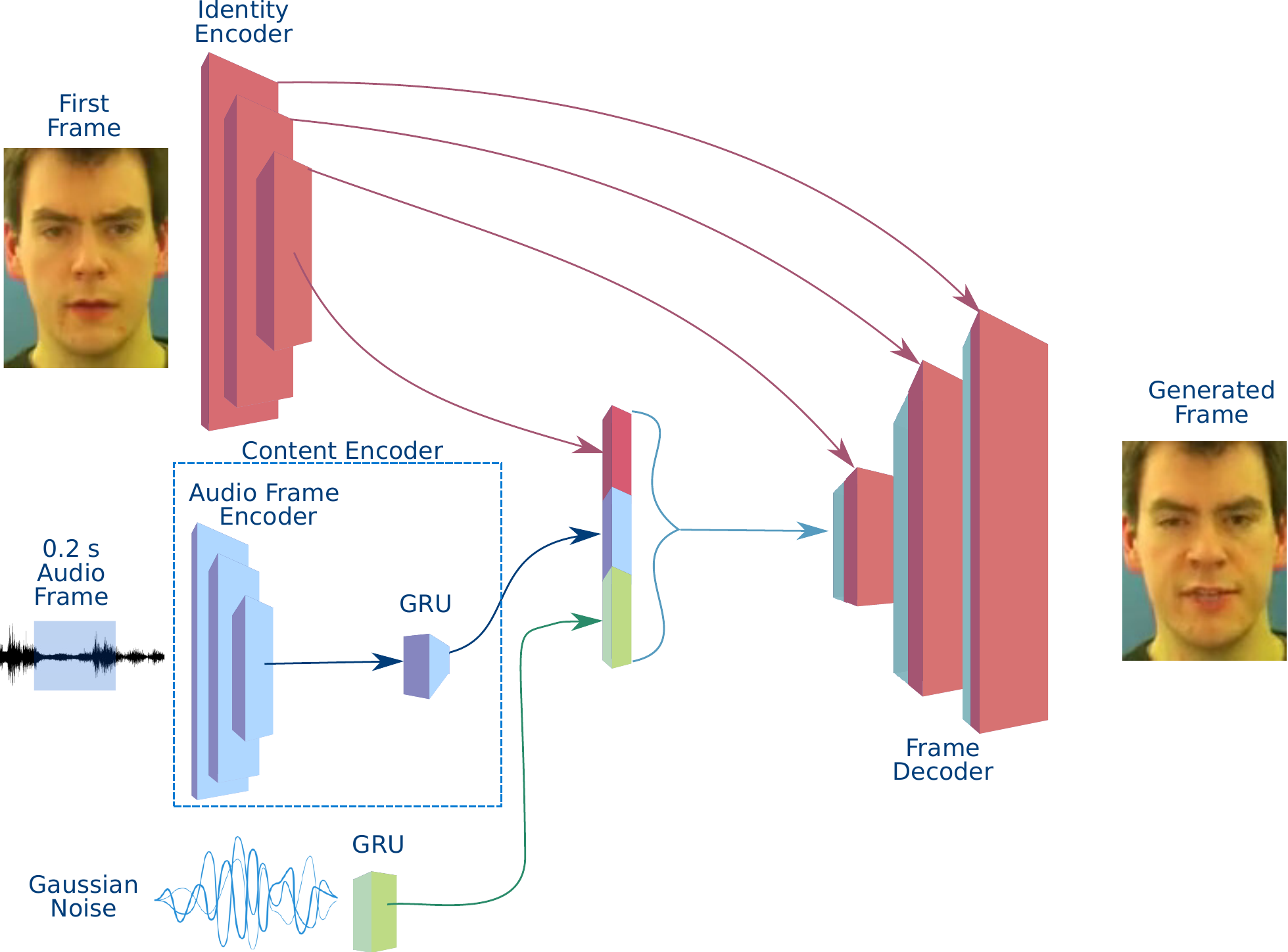}
    \caption{The architecture of the {\em generator} network which consists of a {\em Content Encoder} (audio encoder and RNN), an {\em Identity Encoder}, a {\em Frame Decoder} and {\em Noise Generator}}
    \label{fig:generator_modules}
\end{figure}

The generator network has an encoder-decoder structure and can be conceptually divided into sub-networks as shown in \figref{fig:generator_modules}. We assume a latent representation that is made up of 3 components which account for the speaker identity, audio content and spontaneous facial expressions. These components are generated by different modules and combined to form an embedding which can be transformed into a frame by the decoding network.

\subsubsection{Identity Encoder}
\label{sec:id_encoder}
The speaker's identity is encoded using a 6-layer CNN. Each layer uses strided 2D convolutions, followed by batch normalization and ReLU activation functions. The {\em Identity Encoder} network reduces a $96\times128$ input image to a $128$ dimensional encoding $z_{id}$.

\subsubsection{Content Encoder}
\label{sec:content_encoder}
Audio frames are encoded using a network comprising of 1D convolutions followed by batch normalization and ReLU activation functions. The initial convolutional layer starts with a large kernel, as recommended in \cite{Dai2016}, which helps limit the depth of the network while ensuring that the low-level features are meaningful. Subsequent layers use smaller kernels until an embedding of the desired size is achieved. The audio frame encoding is input into a 1-layer GRU, which produces a content encoding $z_c$ with $256$ elements.

\subsubsection{Noise Generator}
\label{sec:noise_gen}
Although speech contains the necessary information for lip movements it can not be used to produce spontaneous facial expressions. To account for such expressions we propose appending a noise component to our latent representation. Spontaneous expressions such as blinks are coherent facial motions and therefore we expect the latent space that models them to exhibit the same temporal dependency. We therefore, avoid using white noise to model these expressions since it is by definition temporally independent. Instead we use a {\em Noise Generator} capable of producing noise that is temporally coherent. A 10 dimensional vector is sampled from a Gaussian distribution with mean $0$ and variance of $0.6$ and passed through a single-layer GRU to produce the noise sequence. This latent representation introduces randomness in the face synthesis process and helps with the generation of blinks and brow movements.

\subsubsection{Frame Decoder}
\label{sec:frame_decoder}
\begin{figure}[t]
  \centering
  \begin{subfigure}[b]{0.3\linewidth}
    \centering\includegraphics[width=0.6\textwidth]{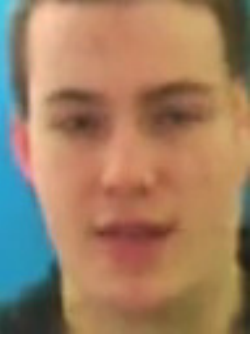}
    \caption{\label{fig:no_skip}}
  \end{subfigure}
  \begin{subfigure}[b]{0.3\linewidth}
    \centering\includegraphics[width=0.6\textwidth]{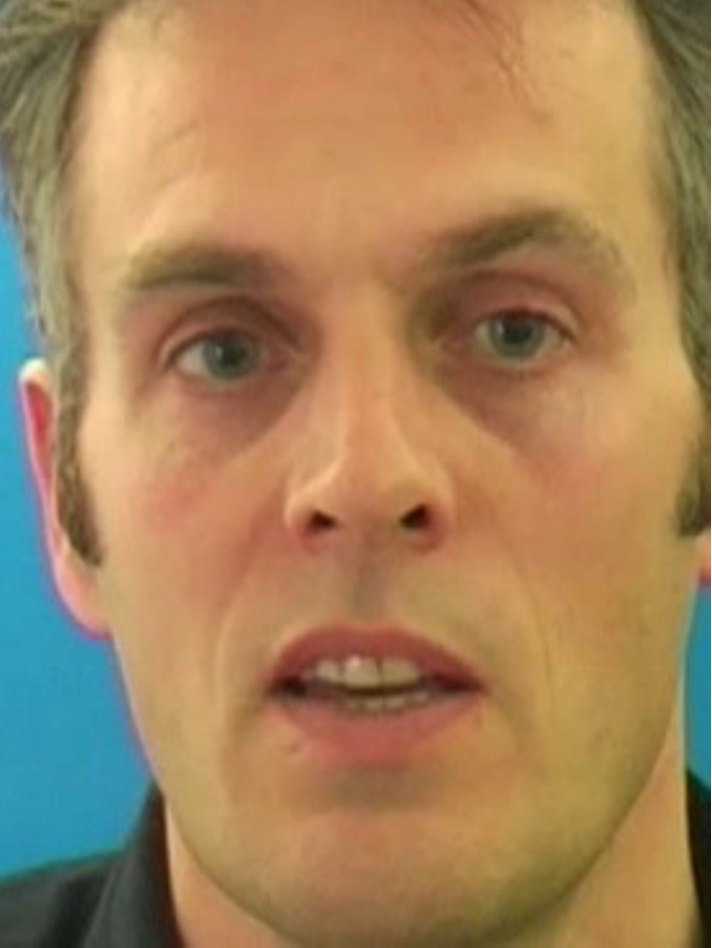}
    \caption{\label{fig:original_frame} }
  \end{subfigure}
  \begin{subfigure}[b]{0.3\linewidth}
    \centering\includegraphics[width=0.6\textwidth]{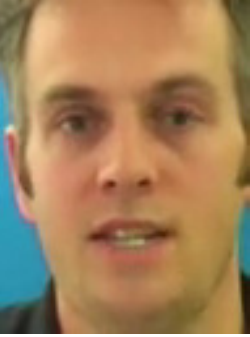}
    \caption{\label{fig:skip connections} }
  \end{subfigure}
\caption{The effect of adding skip connections to the generator network. The frames obtained without skip connections shown in (\subref{fig:no_skip}) do not resemble the person in the ground truth video (\subref{fig:original_frame}). Adding skip connections ensures that the identity is preserved in frames (\subref{fig:skip connections}).}
\label{fig:unet_contribution}
\end{figure}

 The latent representation for each frame is constructed by concatenating the identity, content and noise components. The {\em Frame Decoder} is a CNN that uses strided transposed convolutions to produce the video frames from the latent representation. A U-Net \cite{Ronneberger2015} architecture is used with skip connections between the {\em Identity Encoder} and the {\em Frame Decoder} to preserve the identity of the subject as shown in \figref{fig:unet_contribution}.

\subsection{Discriminators}
\label{sec:discriminators}
Our system uses multiple discriminators in order to capture different aspects of natural videos. The {\em Frame Discriminator} achieves a high-quality reconstruction of the speakers' face throughout the video. The {\em Sequence Discriminator} ensures that the frames form a cohesive video which exhibits natural movements. Finally, the {\em Synchronization Discriminator} reinforces the requirement for audio-visual synchronization.

\subsubsection{Frame Discriminator}
\label{sec:frame_disc}
The {\em Frame Discriminator} is a 6-layer CNN that determines whether a frame is real or not. Adversarial training with this discriminator ensures that the generated frames are realistic. Furthermore, the original still frame is concatenated channel-wise to the target frame and used as a condition, which enforces the identity onto the video frames. 

\subsubsection{Sequence Discriminator}
\label{sec:seq_disc}
The {\em Sequence Discriminator} distinguishes between real and synthetic videos. At every time step the discriminator will use a CNN with spatio-temporal convolutions to extract transient features, which are then fed into a 1-layer GRU. A single layer classifier used at the end of the sequence determines if a sequence is real or not.

\subsubsection{Synchronization Discriminator}
\label{sec:sync_disc}
\begin{figure}[t]
    \centering
    \includegraphics[width=\columnwidth]{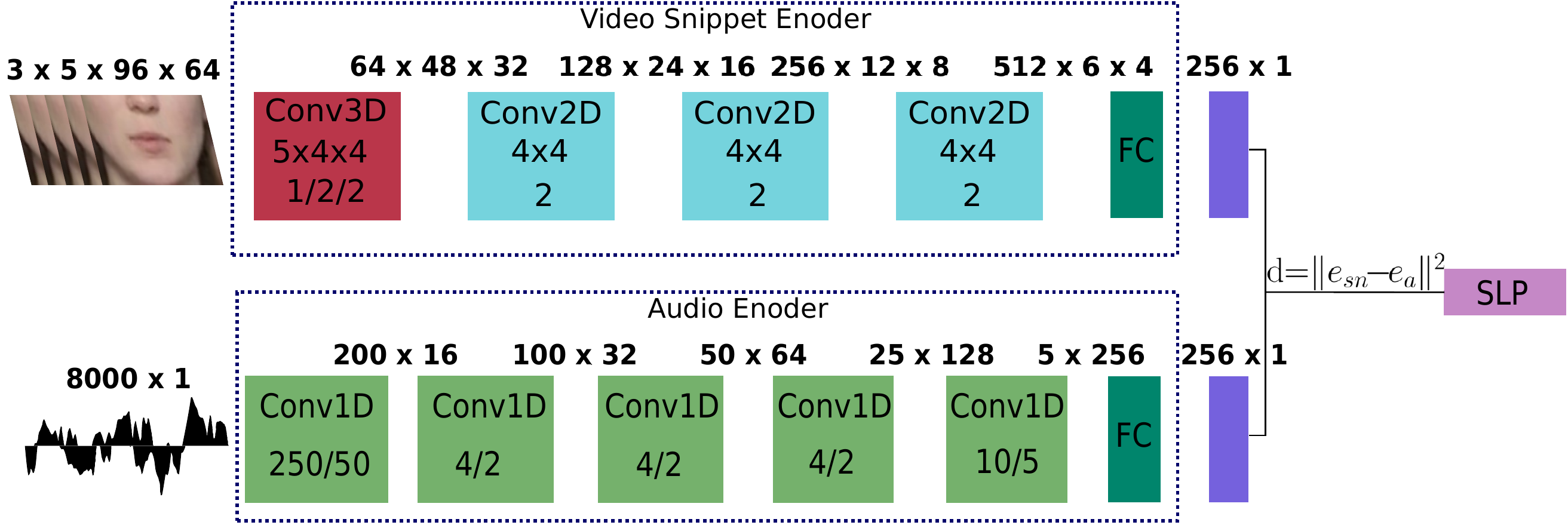}
    \caption{The synchronization discriminator decides if an audio-visual pair is in or out of sync. It uses 2 encoders to obtain embeddings for audio and video and decides if they are in or out of sync based on their Euclidean distance.}
    \label{fig:sync_disc}
\end{figure}

The {\em Synchronization Discriminator} is given fixed-length snippets (corresponding to $0.2s$) of the original video and audio and determines whether they are in or out of sync. This discriminator uses a two stream architecture to compute an embedding for audio and video. The Euclidean distance between the 2 embeddings is calculated and fed into a single layer perceptron for classification. The architecture of this discriminator is shown in \figref{fig:sync_disc}.

% \begin{figure}[h!]
%     \centering
%     \includegraphics[width=\columnwidth]{images/distances.pdf}
%     \caption{The Euclidean distance of the encodings produced by the two streams of the {\em Synchronization Discriminator}. During training the discriminator learns to reduce this distance for synchronised pairs and increase for misaligned pairs.}
%     \label{fig:distances}
% \end{figure}

Showing the discriminator only real or fake audio-video pairs will not necessarily result in samples being classified based on their audio visual correspondence. In order to force the discriminator to judge the sequences based on synchronization we also train it to detect misaligned audio-visual pairs taken from real videos. During training the discriminator learns to reduce the distance between the encodings of synchronized audio-video pairs and increase the distance between misaligned pairs. The distance for the fake pair (generated video with real audio) lies between these two distances and its location is determined by how dominant the discriminator is over the generator. Finally, since movements on the upper half of the face do not affect audio-visual synchrony we have chosen to use only the lower half of the face to train the {\em Synchronization Discriminator}.

\subsection{Training}
\label{sec:training}
\begin{figure}[t]
    \centering
    \includegraphics[width=0.9\columnwidth]{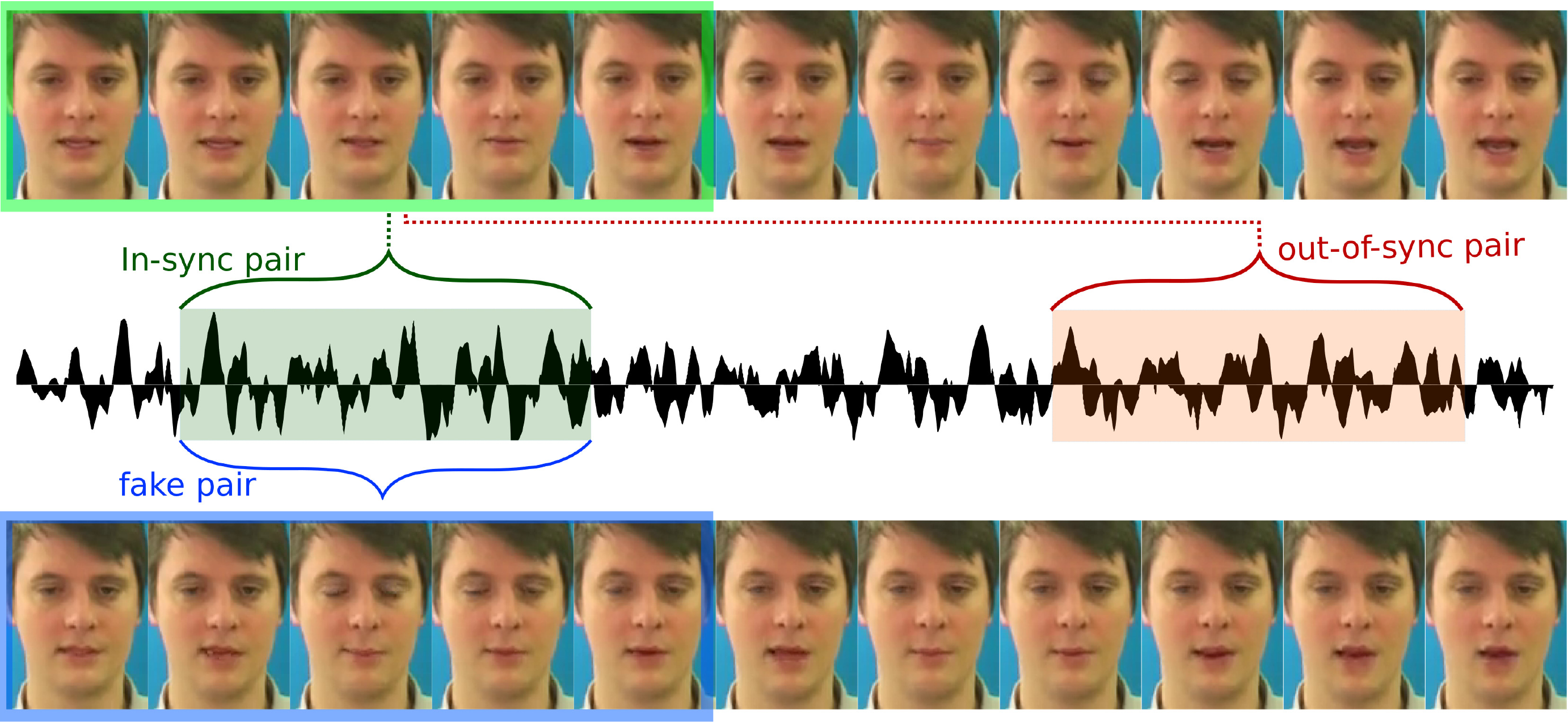}
    \caption{All possible pairs that are used to train the synchronization discriminator. Pairs belong to in one of the following categories \{real video, in-sync audio\}, \{real video, shifted audio\}, \{fake video, matching audio\}}
    \label{fig:snip_selection}
\end{figure}

The {\em Frame discriminator} ($D_{img}$) is trained on frames that are sampled uniformly from a video $x$ using a sampling function $S(x)$. Using the process shown in \figref{fig:snip_selection} we obtain in and out of sync pairs $p_{in}$, $p_{out}$ from the real video $x$ and audio $a$ and a fake pair $p_{f}$.  We use these pairs as training data for the \emph{Synchronization discriminator} ($D_{sync}$). Finally the \emph{Sequence Discriminator} ($D_{seq}$), classifies based on the entire sequence $x$. The total adversarial loss $\pazocal{L}_{adv}$ is made up of the adversarial losses associated with the {\em Frame} ($\pazocal{L}^{img}_{adv}$), {\em Synchronization} ($\pazocal{L}^{sync}_{adv}$) and {\em Sequence} ($\pazocal{L}^{seq}_{adv}$) discriminators. These losses are described by \eref{eq:adv_loss_img} -- \ref{eq:adv_loss_seq}. The total adversarial loss is an aggregate of the losses associated
with each discriminator as shown in \eref{eq:total_adv_loss}, where each loss is assigned a corresponding weight ($\lambda_{img}$, $\lambda_{sync}$, $\lambda_{seq}$).

\begin{small}
\begin{align}
    \begin{split}
        \pazocal{L}^{img}_{adv} = & \EX_{x \sim P_d}[\log D_{img}(S(x), x_1)] + \\ &\EX_{z \sim P_z}[\log (1- D_{img}(S(G(z)), x_1))] 
    \end{split}\label{eq:adv_loss_img}\\
    \begin{split}
        \pazocal{L}^{sync}_{adv} = \; & \EX_{x \sim P_d}[\log D_{sync}(p_{in})] + \frac{1}{2}\EX_{x \sim P_d}[\log 1-D_{sync}(p_{out})] \; +\\
        &\frac{1}{2}\EX_{z \sim P_z}[\log (1- D_{sync}(S_{snip}(p_f))]
    \end{split}\label{eq:adv_loss_sync}\\
    \begin{split}
        \pazocal{L}^{seq}_{adv} = \; & \EX_{x \sim P_d}[\log D_{seq}(x, a)] + \EX_{z \sim P_z}[\log (1- D_{seq}(G(z), a))]
    \end{split}\label{eq:adv_loss_seq}\\
    \begin{split}
        \pazocal{L}_{adv} =\; & \lambda_{img} \pazocal{L}^{img}_{adv} + \lambda_{sync} \pazocal{L}^{sync}_{adv} + \lambda_{seq} \pazocal{L}^{seq}_{adv}
    \end{split}\label{eq:total_adv_loss}
\end{align}
\end{small}

An $L_1$ reconstruction loss is also used to help capture the correct mouth movements. However we only apply the reconstruction loss to the lower half of the image since it discourages the generation of facial expressions. For a ground truth frame $F$ and a generated frame $G$ with dimensions $W \times H$ the reconstruction loss at the pixel level is \eref{eq:reconstruction_loss}.

\begin{equation}
    \pazocal{L}_{L_1} = \sum_{p \in [ 0, W] \times [ \frac{H}{2}, H ] }|F_{p} - G_{p}| 
    \label{eq:reconstruction_loss}
\end{equation}

The loss of our model, shown in \eref{eq:loss}, is made up of the adversarial loss and the reconstruction loss. The $\lambda_{rec}$ hyperparameter controls the contribution of of the reconstruction loss compared to the adversarial loss and is chosen so that, after weighting, this loss is roughly triple the adversarial loss. Through fine tuning on the validation set we find that the optimal values of the loss weights are $\lambda_{rec}= 600$, $\lambda_{img}= 1$, $\lambda_{sync}= 0.8$ and $\lambda_{seq}= 0.2$. The model is trained until no improvement is observed in terms of the audio-visual synchronization on the validation set for 5 epochs. We use pre-trained lipreading models where available or other audio-visual synchronization models to evaluate the audio-visual synchrony of a video. 

\begin{equation}
\begin{split}
\arg \min_{G} \max_{D} \;\;  \pazocal{L}_{adv} +\; \lambda_{rec} \pazocal{L}_{L_1} 
\end{split}
\label{eq:loss}
\end{equation}

We used Adam \cite{Kingma2014} for all the networks with a learning rate of $0.0001$ for the {\em Generator} and {\em Frame Discriminator}. The {\em Sequence Discriminator} and {\em Synchronization Discriminator} use a smaller learning rate of $10^{-5}$. Smaller learning rates for the sequence and synchronization discriminators are required in order to avoid over-training the discriminators, which can lead to instability \cite{Arjovsky2017TowardsNetworks}. The learning rate of the generator and discriminator decays with rates of $2\%$ and $10\%$, respectively, every 10 epochs. 

\section{Datasets}
\label{sec:datasets}
Experiments are run on the GRID, TCD TIMIT, CREMA-D and LRW datasets. The GRID dataset has 33 speakers each uttering 1000 short phrases, containing 6 words randomly chosen from a limited dictionary. The TCD TIMIT dataset has 59 speakers uttering approximately 100 phonetically rich sentences each. Finally, in the CREMA-D dataset 91 actors coming from a variety of different age groups and races utter 12 sentences. Each sentence is acted out by the actors multiple times for different emotions and intensities.

We use the recommended data split for the TCD TIMIT dataset but exclude some of the test speakers and use them as a validation set. For the GRID dataset speakers are divided into training, validation and test sets with a $50\% - 20\% - 30\%$ split respectively. The CREMA-D dataset is also split with ratios $70\% - 15\% - 15\%$ for training, validation and test sets. Finally, for the LRW dataset we use the recommended training, validation and test sets. However we limit our training to faces that are nearly frontal. To do this we use pose estimation software \cite{3ddfa_cleardusk} based on the model proposed in \cite{zhu2017face} to select faces whose roll, pitch and yaw angles are smaller 10\degree.

As part of our pre-processing all faces are aligned to the canonical face and images are normalized. We perform data augmentation on the training set by mirroring the videos. The amount of data used for training and testing is presented in \tabref{tab:Subjects}.

\begin{table}[t]
\small
\begin{center}
\begin{tabular}{|l|l|}
\hline
\multicolumn{1}{|c|}{Dataset} & \multicolumn{1}{c|}{Test Subjects}  \\
\hline\hline
GRID & 2, 4, 11, 13, 15, 18, 19, 25, 31, 33 \\
\hline
TCD TIMIT & 8, 9, 15, 18, 25, 28, 33, 41, 55, 56  \\
\hline
CREMA-D & 15, 20, 21, 30, 33, 52, 62, 81, 82, 89  \\
\hline
\end{tabular}
\end{center}
\caption{The subject IDs that our model is tested on for each dataset.}
\label{tab:test_subjects}
\end{table}

\begin{table}[t]
\small
\begin{center}
\tabcolsep=0.06cm
\begin{tabular}{|l|c|c|c|c|c|}
\hline
Dataset & Samples/Hrs (Tr) & Samples/Hrs (V)&  Samples/Hrs (T) \\
\hline\hline
GRID   & 31639 / 26.4 & 6999 / 5.8 & 9976 / 8.31 \\
TCD    & 8218 / 9.1 & 686 / 0.8 &977 / 1.2 \\
CREMA & 11594 / 9.7 & 819 / 0.7 & 820 / 0.68 \\
LRW & 112658 / 36.3 & 5870 / 1.9& 5980 / 1.9\\
\hline
\end{tabular}
\end{center}
\caption{The samples and hours of video in the training (Tr), validation (V) and test (T) sets.}
\label{tab:Subjects}
\end{table}

\begin{table}[t]
\small
\begin{center}
\begin{tabular}{|c|c|c|c|c|}
\hline
Accuracy & Precision & Recall & MAE (Start) & MAE (End)\\
\hline
80\% & 100\% & 80\% & 1.4 & 2.1\\
\hline
\end{tabular}
\end{center}
\caption{Performance of the blink detector on a small selection of videos from the GRID database that was manually annotated.}
\label{tab:blink-performance}
\end{table}

\section{Metrics}
\label{sec:metrics}
This section describes the metrics that are used to assess the quality of generated videos. The videos are evaluated using traditional image reconstruction and sharpness metrics. Although these metrics can be used to determine frame quality they fail to reflect other important aspects of the video such as audio-visual synchrony and the realism of facial expressions. We therefore propose using alternative methods that are capable of capturing these aspects of the generated videos.
\begin{description}[leftmargin=0cm]
\item [\bt{Reconstruction Metrics}:]

We use common reconstruction metrics such as the peak signal-to-noise ratio (PSNR) and the structural similarity (SSIM) index to evaluate the generated videos. During our assessment it is important to take into account the fact that reconstruction metrics will penalize videos for any facial expression that does not match those in the ground truth videos. 

\item [\bt{Sharpness Metrics}:]
The frame sharpness is evaluated using the cumulative probability blur detection (CPBD) measure \cite{Narvekar2009}, which determines blur based on the presence of edges in the image. For this metric as well as for the reconstruction metrics larger values imply better quality.

\item [\bt{Content Metrics}:]
The content of the videos is evaluated based on how well the video captures identity of the target and on the accuracy of the spoken words. We verify the identity of the speaker using the average content distance (ACD) \cite{Tulyakov2017}, which measures the average Euclidean distance of the still image representation, obtained using OpenFace \cite{Amos2016}, from the representation of the generated frames. The accuracy of the spoken message is measured using the word error rate (WER) achieved by a pre-trained lip-reading model. We use the LipNet model \cite{Assael2016}, which surpasses the performance of human lip-readers on the GRID dataset. For both content metrics lower values indicate better accuracy.

\item [\bt{Audio-Visual Synchrony Metrics}:]
 Synchrony is quantified using the methods proposed in \cite{Chung2016OutWild}. In this work Chung \etal propose the SyncNet network which calculates the euclidean distance between the audio and video encodings on small (0.2 second) sections of the video. The audio-visual offset is obtained by using a sliding window approach to find where the distance is minimized. The offset is measured in frames and is positive when the audio leads the video. For audio and video pairs that correspond to the same content the distance will increase on either side of point where the minimum distance occurs. However, for uncorrelated audio and video the distance is expected to be stable. Based on this fluctuation Chung \etal \cite{Chung2016OutWild} further propose using the difference between the minimum and the median of the Euclidean distances as an audio-visual (AV) confidence score which determines the audio-visual correlation. Higher scores indicate a stronger correlation, whereas confidence scores smaller than 0.5 indicate that audio and video are uncorrelated.

\begin{figure}[t]
  \centering
  \begin{subfigure}[b]{0.37\linewidth}
    \centering\includegraphics[width=0.99\textwidth]{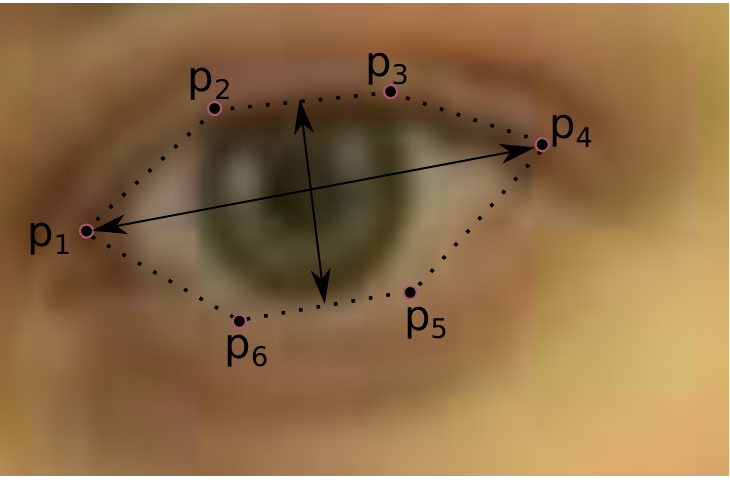}
    \caption{\label{fig:ear_open}}
  \end{subfigure}
  \begin{subfigure}[b]{0.37\linewidth}
    \centering\includegraphics[width=0.99\textwidth]{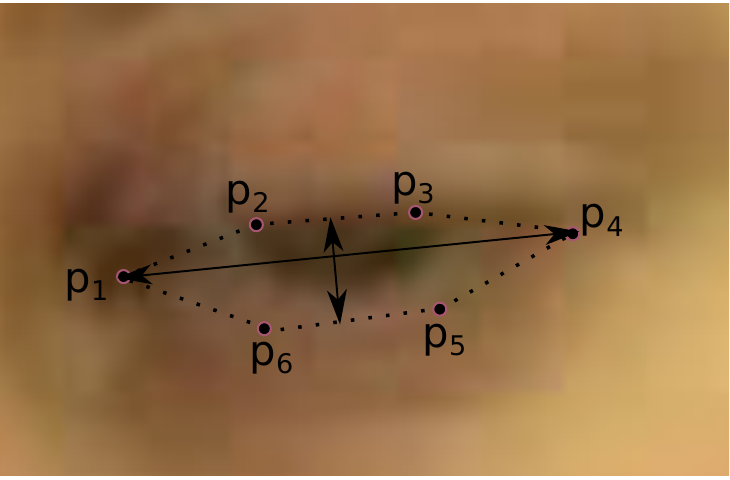}
    \caption{\label{fig:ear_closed} }
  \end{subfigure}
\caption{Landmarks used for EAR calculation. An open eye will have a larger EAR compared to a closed eye.}
\label{fig:ear}
\end{figure}

\begin{figure}[t]
    \includegraphics[width=\columnwidth]{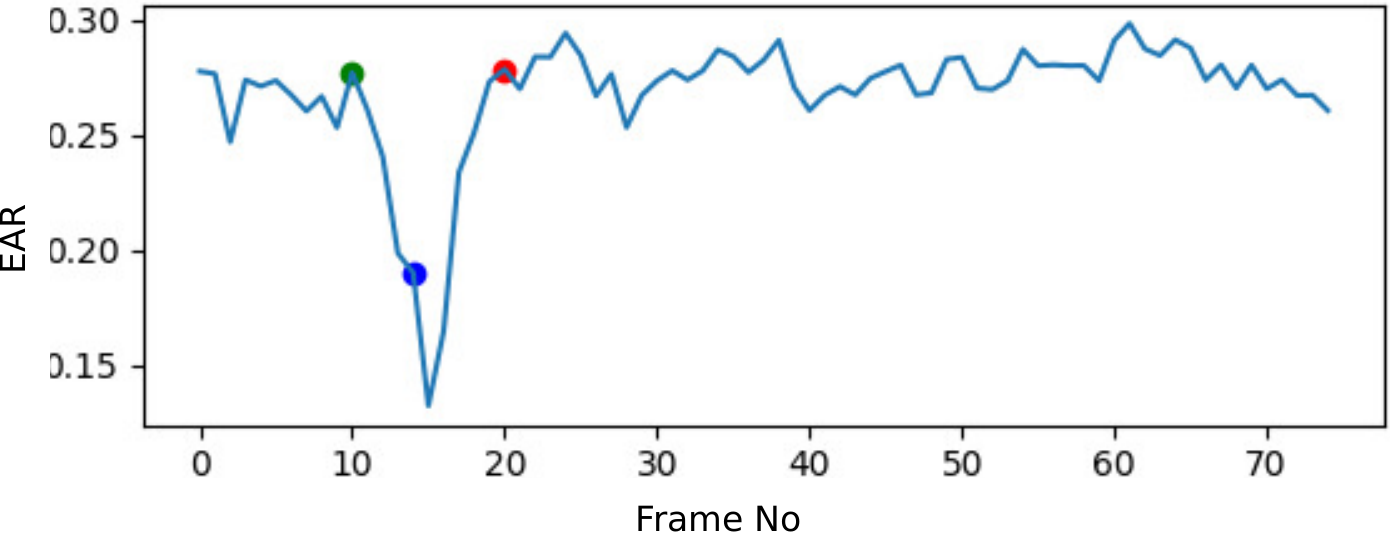}
    \caption{A blink is detected at the location where a sharp drop occurs in the EAR signal (blue dot). We consider the start (green dot) and end (red dot) of the blink to correspond to the peaks on either side of the blink location.}
    \label{fig:5}
\end{figure}

\item [\bt{Expression Evaluation}:]
We investigate the generation of spontaneous expressions since it is one of the main factors that affect our perception of how natural a video looks. According to the study presented in \cite{Bentivoglio1997AnalysisSubjects} the average person blinks 17 times per minute (0.28 blinks/sec), although this rate increases during conversation and decreases when reading. We use a blink detector based on the one proposed in \cite{Soukupova2016Real-timeLandmarks}, which relies on the eye aspect ratio (EAR) to detect the occurrence of blinks in videos. The EAR is calculated per frame according to the formula shown in eq. \eqref{eq:EAR} using facial landmarks $p_1$ to $p_6$ shown in \figref{fig:ear}. The blink detector algorithm first calculates the EAR signal for the entire video and then identifies blink locations by detecting a sharp drop in the EAR signal.

\end{description}

\begin{equation}
    EAR =\frac{ \| p_2-p_6\| + \| p_3-p_5\|}{\| p_1-p_4\|}
    \label{eq:EAR}
\end{equation}

Once the blink is detected we can identify the start and end of the blink by searching for the peaks on either side of that location as shown in \figref{fig:5}. Using this information we can calculate the duration of blinks and visualize the blink distribution.

To gauge the performance of the blink detector we measure its accuracy on 50 randomly selected videos from the GRID validation set that we have manually annotated. The performance metrics for the blink detection as well as the mean absolute error (MAE) for detecting the start and end points of the blinks are shown in \tabref{tab:blink-performance}. The MAE is measured in frames and the video frame rate is 25 fps.

This method detects blinks with a high accuracy of 80\%, which means that we can rely on it to give us accurate statistics for the generated videos. We have chosen a very strict threshold for the drop in EAR in order to ensure that there are minimal if any false alarms. This is evident by the very high precision of the method. Finally, we note that the detector detects the start and end of a blink with an average error of $1.75$ frames.

\begin{figure}[t]
  \centering
  \begin{subfigure}[b]{0.45\linewidth}
    \centering\includegraphics[width=0.99\textwidth]{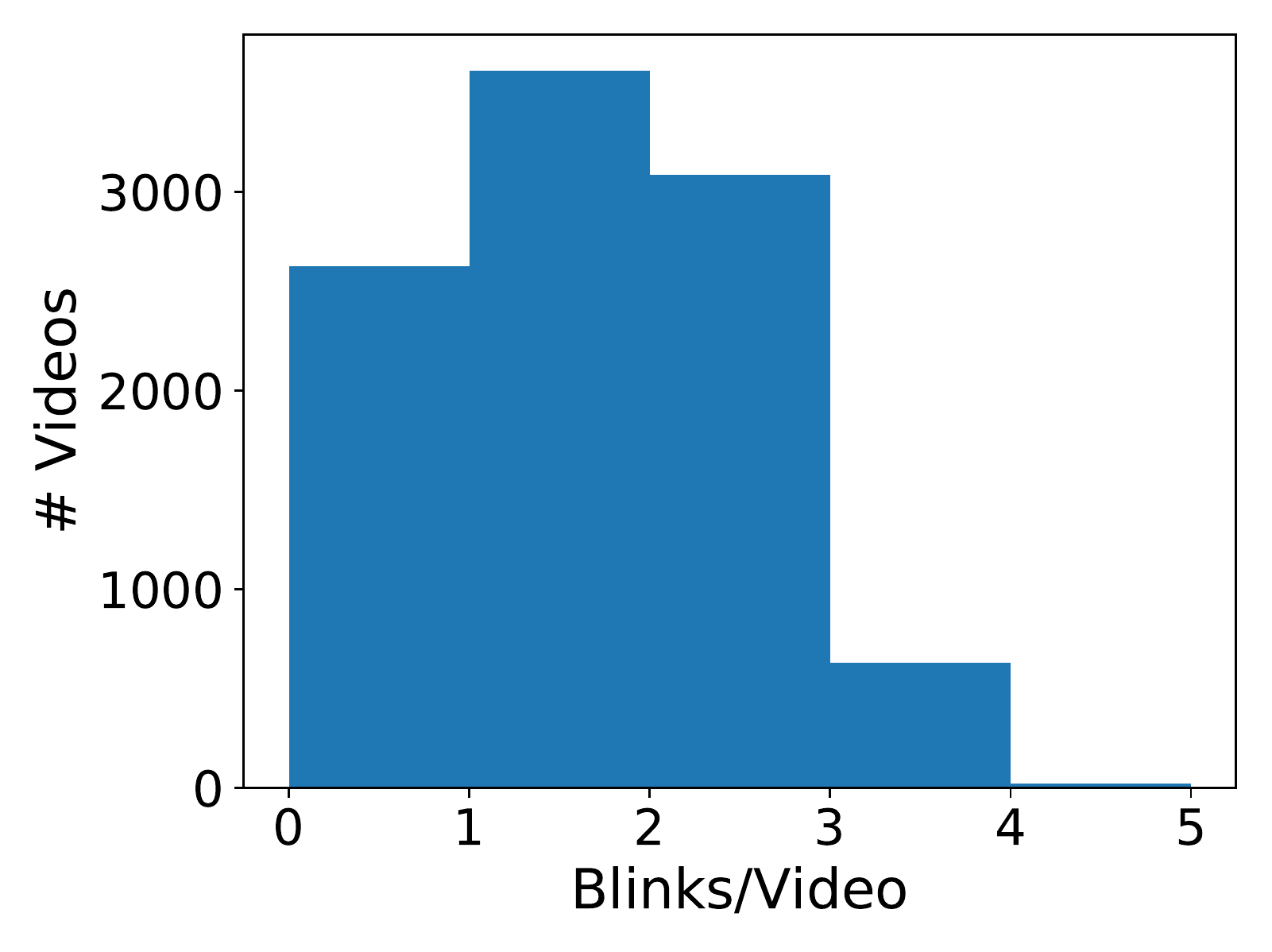}
    \caption{\label{fig:blink_distro}}
  \end{subfigure}
  \begin{subfigure}[b]{0.45\linewidth}
    \centering\includegraphics[width=0.99\textwidth]{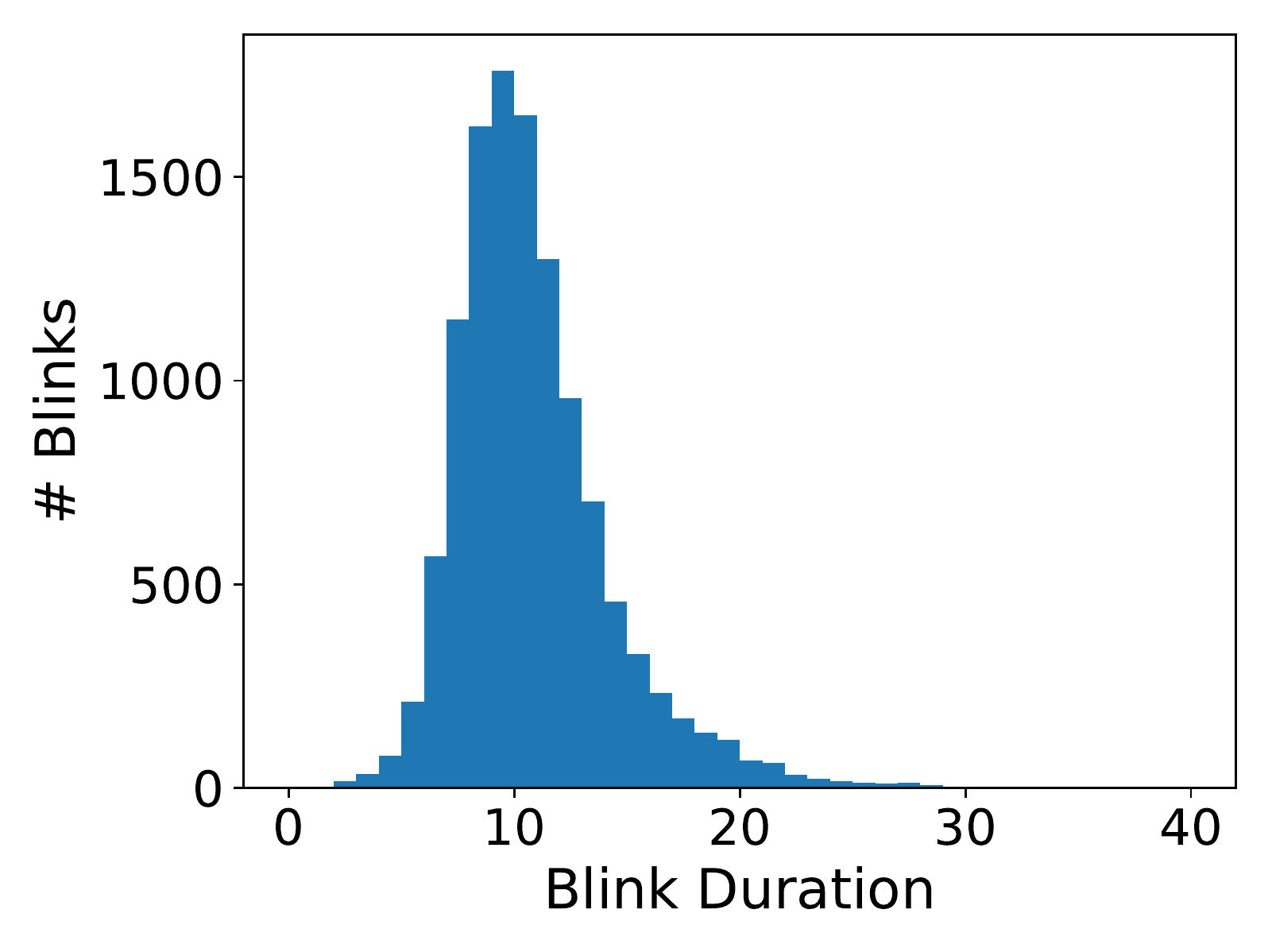}
    \caption{\label{fig:blink_duration_distro} }
  \end{subfigure}
\caption{The distributions for (\subref{fig:blink_distro}) amount of blinks per video and (\subref{fig:blink_duration_distro}) the average blink duration per video from the GRID dataset.}
\label{fig:real_blink}
\end{figure}

We can use the blink detector to obtain the distribution for the number of blinks per video (GRID videos are 3 seconds long) as well as the distribution for blink duration for the GRID test set.  These results are shown in  \figref{fig:real_blink}. The mean blink rate is 1.18 blinks/video or 0.39 blinks/second  which is similar to the average human blink rate of 0.28 blinks/second, especially when considering that the blink rate increases to 0.4 blinks/second during conversation. The average duration of a blink was found to be 10 frames (0.41s). However, we find that using the median is more accurate since this is less sensitive to outliers caused by the detector missing the end of the blink. Finally, it is important to note that the short length of the videos will affect our estimate of the blink rate. The blinks for all the datasets are shown in \tabref{tab:blink_dataset}.

\begin{table}[t]
\small
\begin{center}
\begin{tabular}{|c|c|c|c|c|}
\hline
& GRID & TIMIT & CREMA & LRW\\
\hline
blinks/sec& 0.39 & 0.28 & 0.26 & 0.53\\
median duration (sec)& 0.4 & 0.2& 0.36 & 0.32\\
\hline
\end{tabular}
\end{center}
\caption{The average blink rate and median blink duration for real videos in each dataset.}
\label{tab:blink_dataset}
\end{table}

\section{Experiments}
\label{sec:experiments}
Our model is implemented in PyTorch  and takes approximately a week to train using a single Nvidia GeForce GTX 1080 Ti GPU. During inference the average generation time per frame is 7ms on the GPU, permitting the use of our method in real time applications. A sequence of 75 frames can be synthesized in 0.5s. The frame and sequence generation times increase to 1s and 15s respectively when processing is done on the CPU.

\subsection{Ablation Study}
\label{sec:ablation}
In order to quantify the effect of each component of our system we perform an ablation study on the GRID dataset (see \tabref{tab:ablation}). We use the metrics from section \ref{sec:metrics} and a pre-trained LipNet model which achieves a WER of $21.76 \%$ on the ground truth videos. The average value of the ACD for ground truth videos of the same person is $0.98 \cdot 10^{-4}$ whereas for different speakers it is $1.4 \cdot 10^{-3}$.

\begin{table*}[ht]
\small
\begin{center}
\begin{tabular}{|l|c|c|c|c|c|c|c|c|c|}
\hline
\multicolumn{1}{|l|}{Method} & \multicolumn{1}{c|}{PSNR} & \multicolumn{1}{c|}{SSIM} & \multicolumn{1}{c|}{CPBD} & \multicolumn{1}{c|}{ACD} & \multicolumn{1}{c|}{WER} & \multicolumn{1}{c|}{AV Offset} & \multicolumn{1}{c|}{AV Confidence} & \multicolumn{1}{c|}{blinks/sec}& \multicolumn{1}{c|}{blink dur. (sec)}\\
\hline\hline
GT & $\infty$ & 1.00 & 0.276 & $0.98 \cdot 10^{-4}$ & 21.76\% & 1 & 7.0 & 0.39 & 0.41\\
w/o $\pazocal{L}_{adv}$ & \bt{28.467}  & \bt{0.855} &0.210 & $ 1.92 \cdot 10^{-4}$ & 26.6\% & 1 &7.1 & 0.02 & 0.16\\
w/o $\pazocal{L}_{L_1}$ & 26.516 & 0.805 &\bt{0.270} & $ \bt{1.03} \cdot 10^{-4}$ & 56.4\% & 1 & 6.3 & 0.41 & 0.32\\
w/o $\pazocal{L}^{img}_{adv}$ & 26.474 & 0.804 &0.252 & $1.96 \cdot 10^{-4}$ & 23.2\% & 1 & 7.3 & 0.16 & 0.28\\
w/o $\pazocal{L}^{sync}_{adv}$  & 27.548 & 0.829 & 0.263 & 1.19 $\cdot 10^{-4}$ & 27.8\% & 1  & 7.2 & 0.21 & 0.32\\
w/o $\pazocal{L}^{seq}_{adv}$ & 27.590  & 0.829 & 0.259 & 1.13 $\cdot 10^{-4}$  & 27.0\% & 1 & \bt{7.4} & 0.03 & 0.16\\
Full Model & 27.100 & 0.818 &0.268 & $ 1.47 \cdot 10^{-4}$ & \bt{23.1\%} & 1 & \bt{7.4} & 0.45 & 0.36\\
\hline
\end{tabular} 
\end{center}
\centering
\caption{Ablation study performed on the GRID dataset. In every experiment we train the model by removing a single term from eq. \eqref{eq:loss}.}
\label{tab:ablation}
\end{table*}

\begin{figure}[t]
  \centering
  \begin{subfigure}[b]{0.8\linewidth}
    \centering\includegraphics[width=0.99\textwidth]{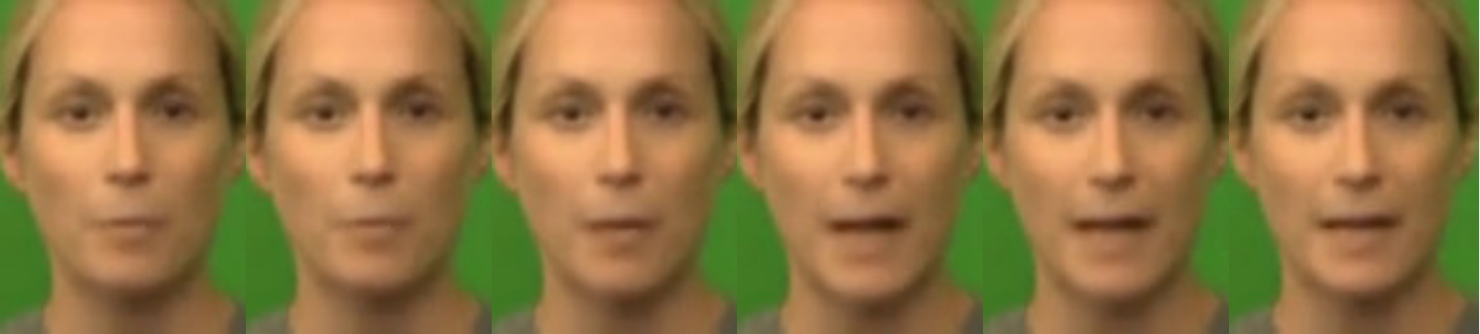}
    \caption{\label{fig:bluriness_l1} $L_1$ loss on entire frame}
  \end{subfigure}
  \begin{subfigure}[b]{0.8\linewidth}
    \centering\includegraphics[width=0.99\textwidth]{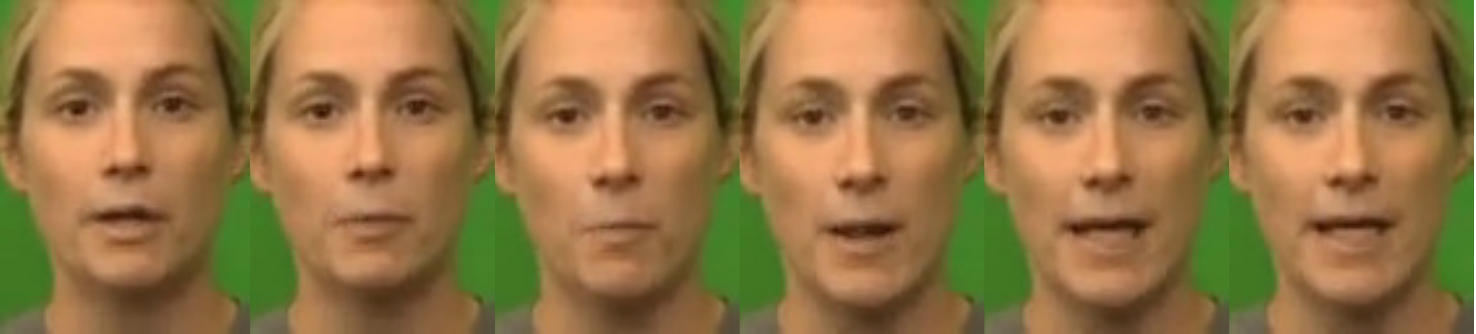}
    \caption{\label{fig:bluriness_model}  Proposed loss on frames}
  \end{subfigure}
  \caption{Frames using (\subref{fig:bluriness_l1}) only an $L_1$ loss on the entire face compared to (\subref{fig:bluriness_model}) frames produced using the proposed method. Frames are taken from videos generated on the CREMA-D test set.}
\label{fig:bluriness}
\end{figure}

The model that uses only an $L_1$ loss achieves better PSNR and SSIM results, which is expected as it does not generate spontaneous expressions, which are penalized by these metrics unless they happen to coincide with those in ground truth videos. We also notice that it results in the most blurry images. The blurriness is minimized when using the frame adversarial loss as indicated by the higher CPBD scores. This is also evident when comparing video frames generated with and without adversarial training as shown in \figref{fig:bluriness}.

\begin{figure}[t]
    \centering
    \includegraphics[width=0.9\columnwidth]{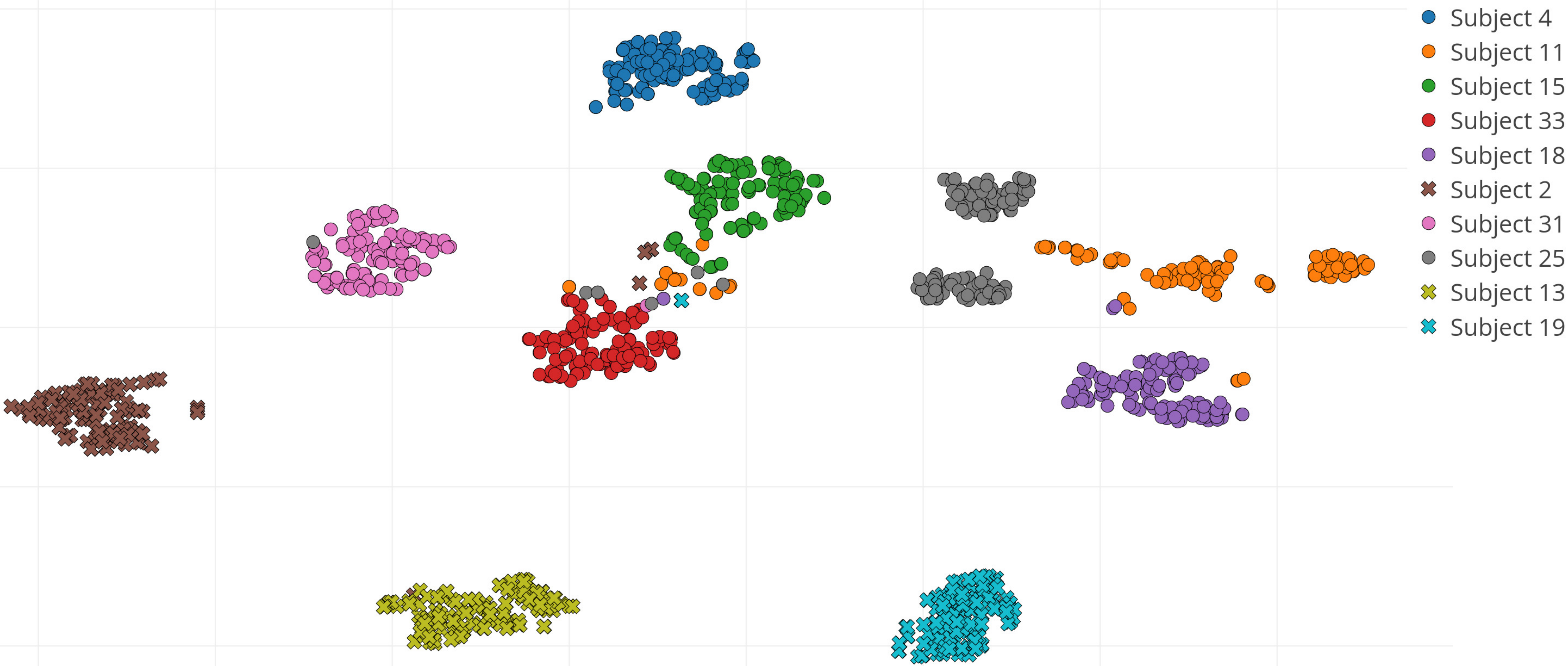}
    \caption{t-SNE plot of the identity encoding of random frames from the GRID test set. Frames corresponding to the same subject have the same colour. Male subjects are indicated by a cross whereas female subjects are indicated by a circle.}
    \label{fig:tsne}
\end{figure}

The Average Content Distance is close to that of the real videos, showing that our model captures and maintains the subject identity throughout the video. Based on the results of the ablation study this is in large part due to the {\em Frame Discriminator}. Furthermore, this indicates that the identity encoder has managed to capture the speaker identity. Indeed, when plotting the identity encoding (\figref{fig:tsne}) of 1250 random images taken from the GRID test set using the t-Distributed Stochastic Neighbor Embedding (t-SNE) algorithm \cite{VanDerMaaten2008VisualizingT-sne} we notice that images of the same subject have neighbouring encodings. Additionally, we notice that the data points can be separated according to gender.

\begin{figure}[t]
  \centering
  \begin{subfigure}[b]{0.45\linewidth}
    \centering\includegraphics[width=\textwidth]{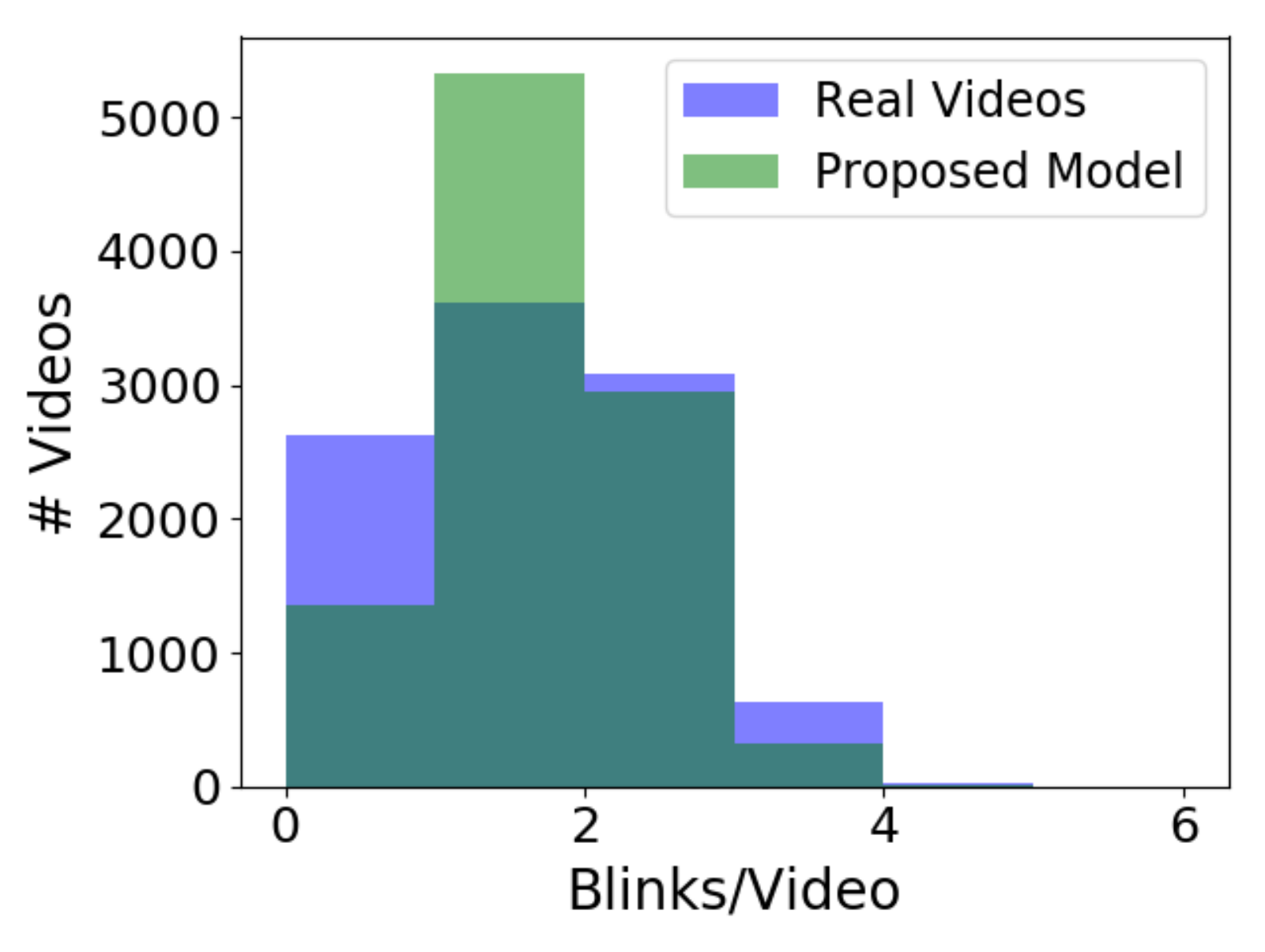}
    \caption{\label{fig:real_full_blinks}  Full model}
  \end{subfigure}%
  \begin{subfigure}[b]{0.45\linewidth}
    \centering\includegraphics[width=\textwidth]{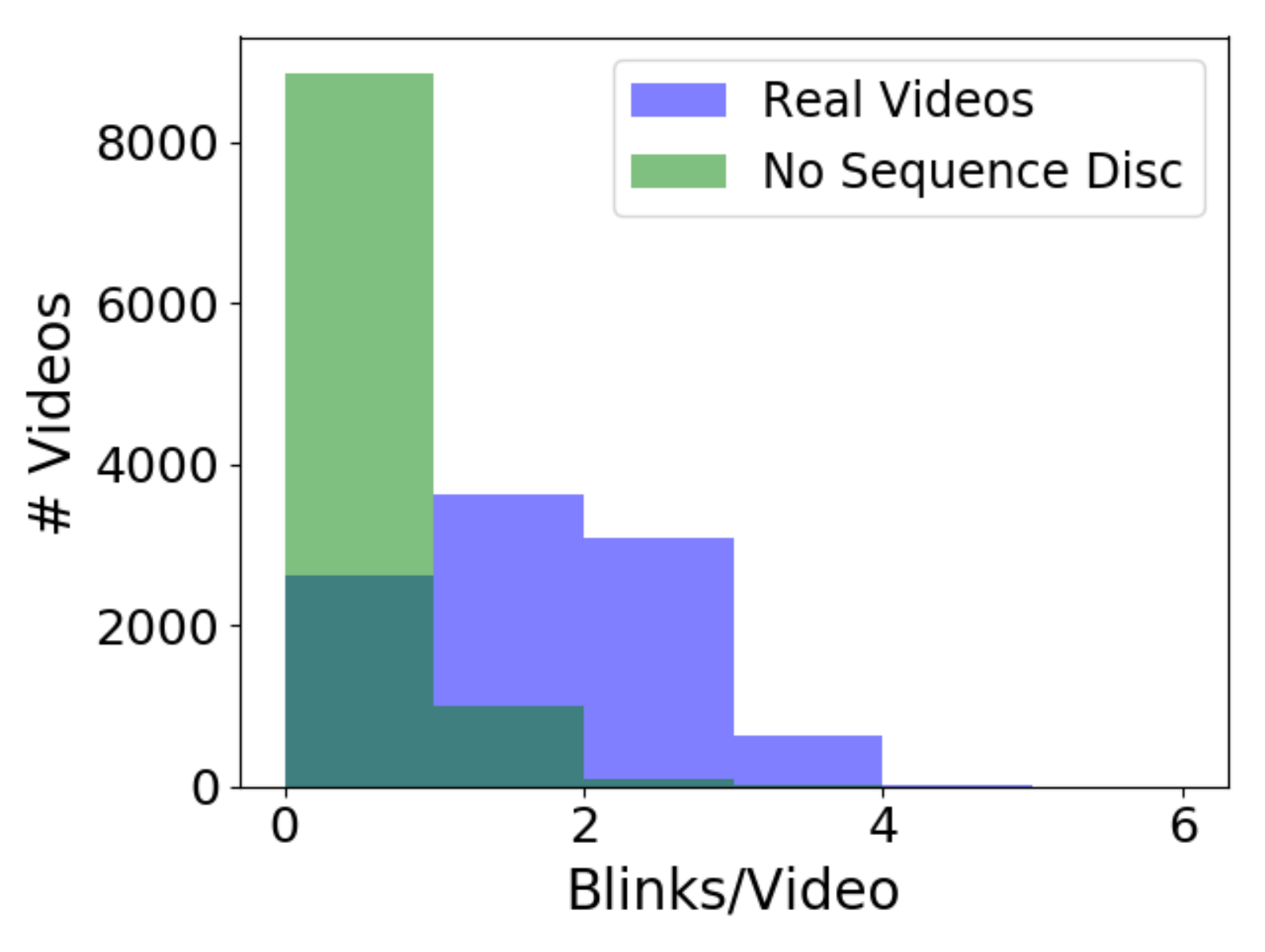}
    \caption{\label{fig:real_noseq_blinks} w/o $\pazocal{L}^{seq}_{adv}$ }
  \end{subfigure}
  \begin{subfigure}[b]{0.45\linewidth}
    \centering\includegraphics[width=\textwidth]{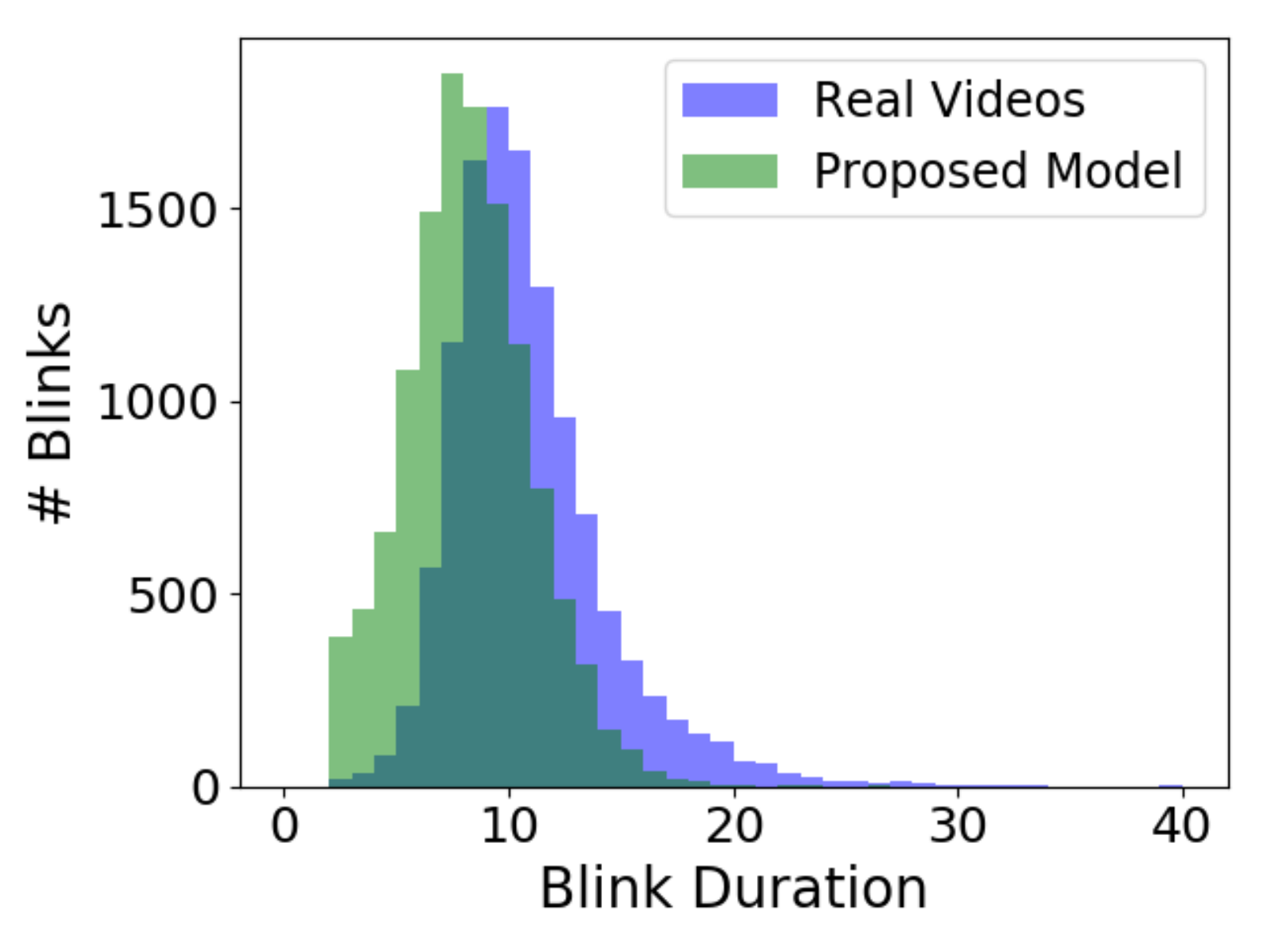}%
    \caption{\label{fig:real_full_durations} Full model}
  \end{subfigure}
  \begin{subfigure}[b]{0.45\linewidth}
    \centering\includegraphics[width=\textwidth]{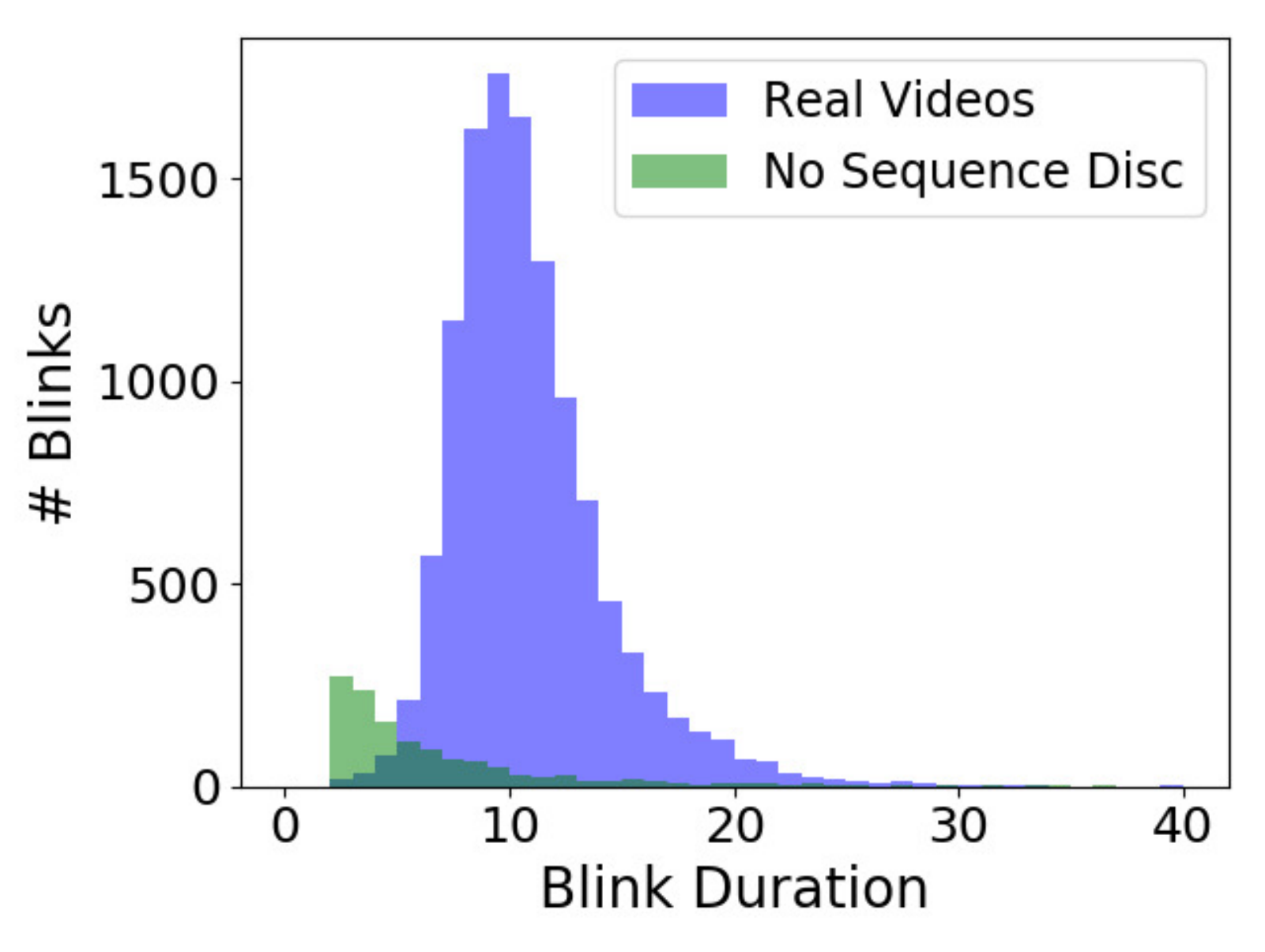}
    \caption{\label{fig:real_noseq_durations} w/o $\pazocal{L}^{seq}_{adv}$}
  \end{subfigure}
  \caption{The distribution of blinks for videos generated by (\subref{fig:real_full_blinks}) our proposed model and (\subref{fig:real_noseq_blinks}) a model without the {\em Sequence Discriminator}. When the {\em Sequence Discriminator} is used (\subref{fig:real_full_durations}) the distribution of blink duration closely resembles that of the real videos. The same does not hold when (\subref{fig:real_noseq_durations}) the {\em Sequence Discriminator} is omitted.}
\label{fig:blink_distro_comp}
\end{figure}

The \emph{Sequence Discriminator} is responsible for the generation of natural expressions. To quantify its effect we compare the distribution of blinks for videos generated by the full model to those generated without the \emph{Sequence Discriminator}. This is shown in \figref{fig:blink_distro_comp},  where it is evident that removing the sequence discriminator drastically reduces blink generation. Furthermore, we note the similarity of the generated and real distribution of blinks and blink duration. The average blink rate in videos generated by our model is 0.4 blinks/sec with the meadian blink lasting 9 frames (0.36s). Both the average blink rate and median duration are very close to those found in the ground truth videos in \tabref{tab:blink_dataset}.

We also notice that the removal of the sequence discriminator coincides with a an increase in PSNR and SSIM, which is likely due to the generation of blinks and head movements. We test this hypothesis by calculating the PSNR only on the lower half of the image and find that gap between the non-adversarial model and our proposed model reduces by 0.3 dB.

The effect of the synchronization discriminator is reflected in the low WER and high AV confidence values. Our ablation study shows that the temporal discriminators have a positive contribution to both the audio-visual synchronization and the WER.

% We can compare the facial movements of a video generated by our model to those of a real video by plotting the optical flow motion map. Motion maps are images in the Hue Saturation Value (HSV) color-space, where hue reflects the angle of movement and the value reflects the magnitude of the motion. These images are obtained by calculating the optical flow between consecutive frames. An example for real and generated videos is shown in \figref{fig:heatmap}.

% \begin{figure}[t!]
%   \centering
%   \begin{subfigure}[b]{0.9\linewidth}
%     \centering\includegraphics[width=0.99\textwidth]{images/heatmap_real.pdf}
%     \caption{\label{fig:heatmap_real}}
%   \end{subfigure}
%   \begin{subfigure}[b]{0.9\linewidth}
%     \centering\includegraphics[width=0.99\textwidth]{images/heatmap_model.pdf}
%     \caption{\label{fig:heatmap_gen} }
%   \end{subfigure}
% \caption{Motion maps created using optical flow for (\subref{fig:heatmap_real}) real video and (\subref{fig:heatmap_gen}) video generated using our best model. Reference of motion is shown on the top left. Red and green colors correspond to upwards and downwards motion respectively.}
% \label{fig:heatmap}
% \end{figure}

\begin{figure*}[t]
\begin{center}
\includegraphics[width=0.99\linewidth]{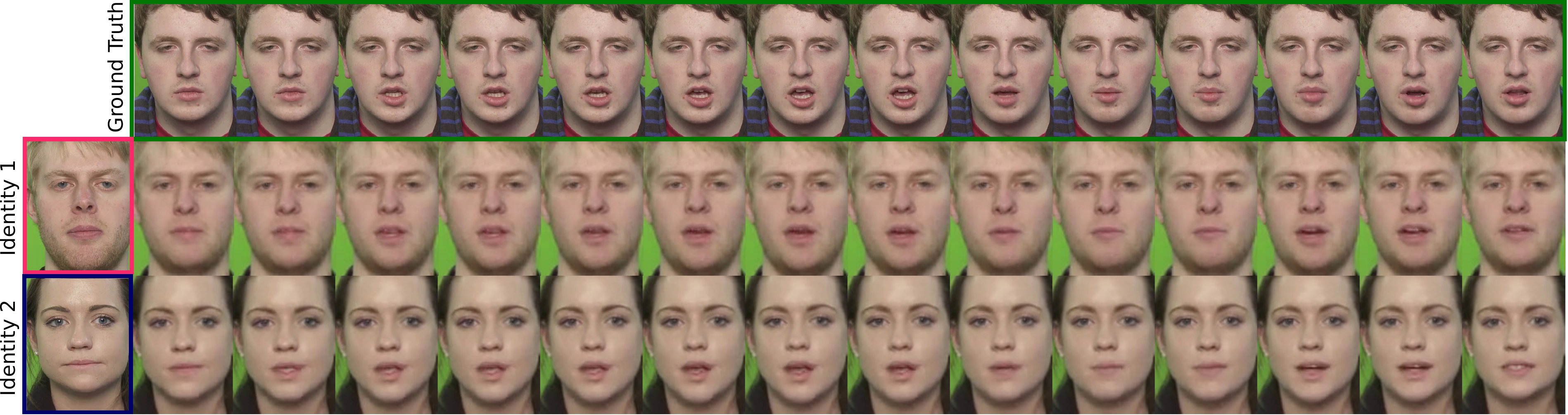}\\
\end{center}
\caption{Animation of different faces using the same audio. The movement of the mouth is similar for both faces as well as for the ground truth sequence. Both audio and still image are taken from the TIMIT dataset and are unseen during training.}
\label{fig:diffface}
\end{figure*}

%\begin{figure}[t]
%  \centering
%  \begin{subfigure}[b]{0.95\linewidth}
%    \centering\includegraphics[width=0.99\textwidth]{images/bin_female.pdf}
%    \caption{\label{fig:diff_audioa} Female voice uttering the word ``bin''}
   
%  \end{subfigure}
 % \begin{subfigure}[b]{0.95\linewidth}
%    \centering\includegraphics[width=0.99\textwidth]{images/white_male.pdf}
%    \caption{\label{fig:diff_audiob} Male voice uttering the word ``white''}
%  \end{subfigure}
%\caption{Generated sequences for %(\subref{fig:diff_audioa}) the word ``bin'' %(\subref{fig:diff_audiob}) the word ``white'' from the GRID test set. Coarticulation is evident in (\subref{fig:diff_audioa}) where ``bin'' is followed by the word ``blue''.}
%\label{fig:diff_audio}
%\end{figure}

\begin{figure}[t]
  \centering
  \begin{subfigure}[b]{\linewidth}
    \centering\includegraphics[width=0.08\textwidth]{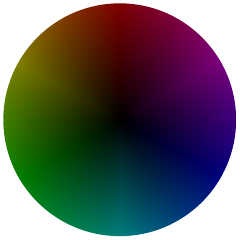}
    \caption{\label{fig:reference} Movement direction map}
  \end{subfigure}
  \begin{subfigure}[b]{\linewidth}
    \centering\includegraphics[width=0.95\textwidth]{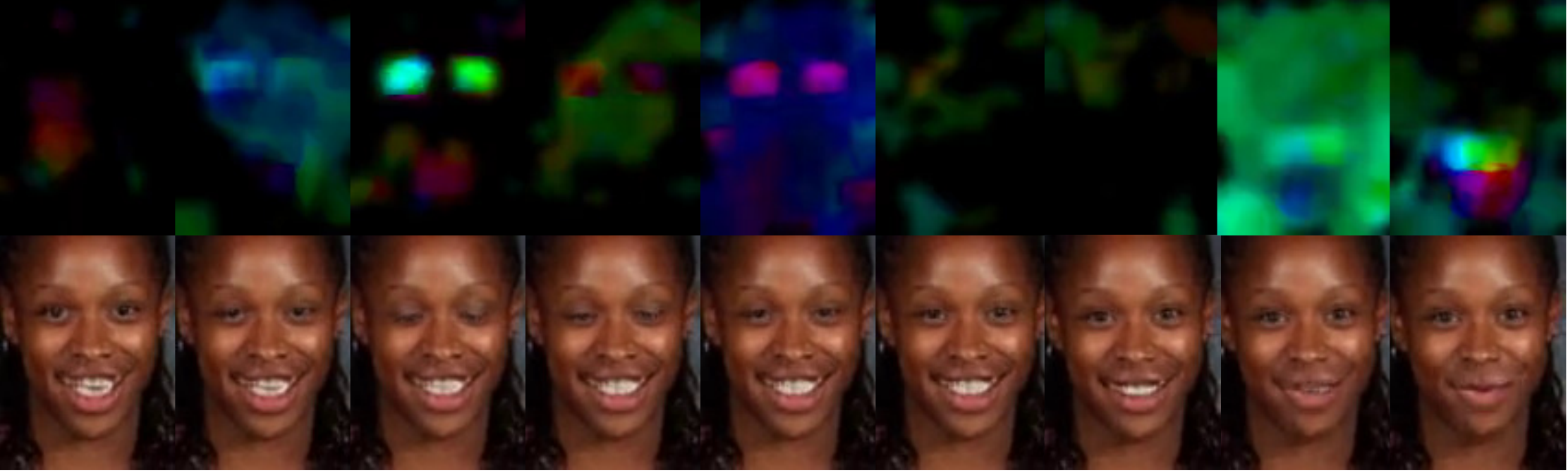}
    \caption{\label{fig:blink} Generated blink using audio from LRW and image from CelebA}
  \end{subfigure}
  \begin{subfigure}[b]{\linewidth}
    \centering\includegraphics[width=0.95\textwidth]{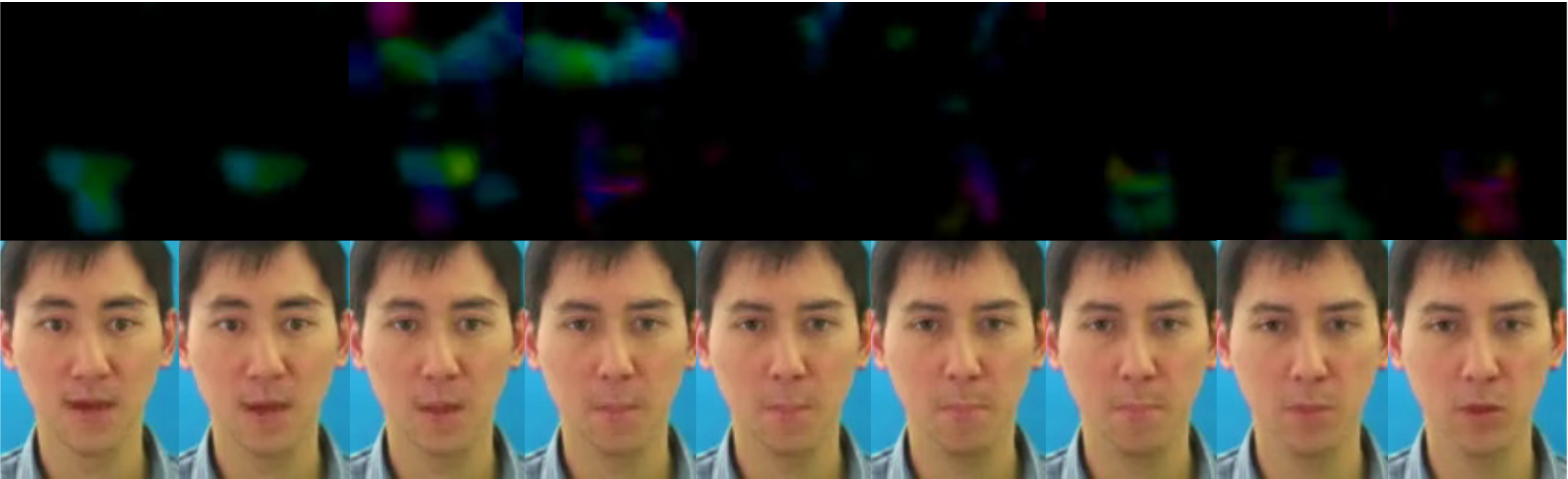}
    \caption{\label{fig:frown} Generated frown on GRID dataset}
  \end{subfigure}
  \begin{subfigure}[b]{\linewidth}
    \centering\includegraphics[width=0.95\textwidth]{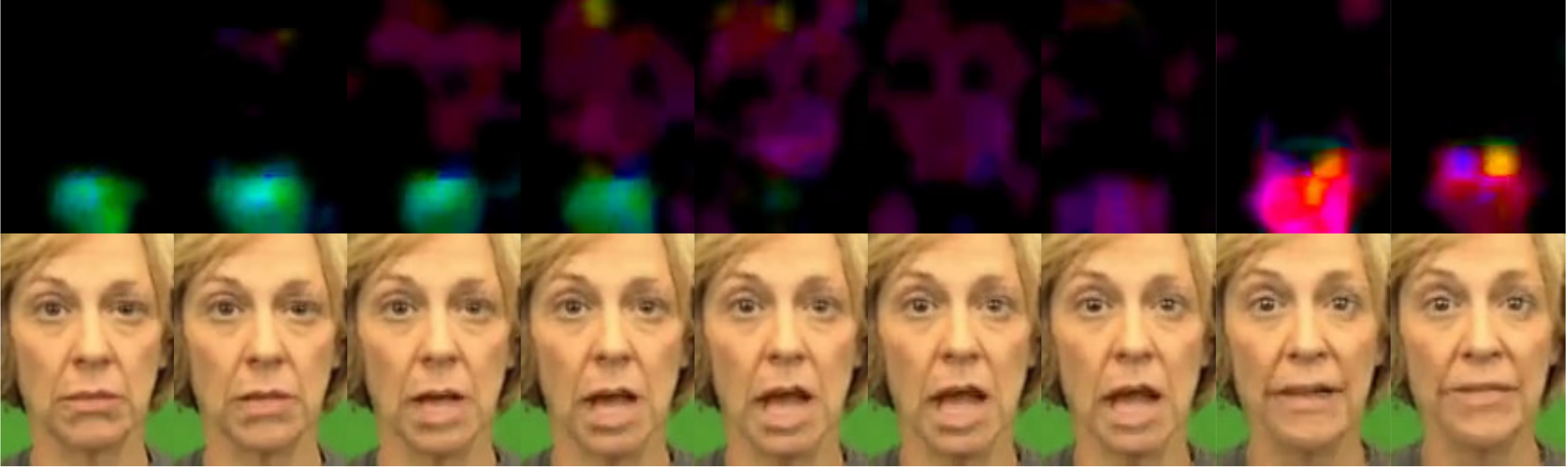}
    \caption{\label{fig:shouting} Angry expression from shouting audio on CREMA-D dataset}
  \end{subfigure}
  \caption{Facial expressions generated using our framework include (\subref{fig:blink}) blinks, (\subref{fig:frown}) frowns and (\subref{fig:shouting}) shouting expressions. The corresponding optical flow motion map is shown above each sequence. A reference diagram for the direction of the movement is shown in (\subref{fig:reference}).}
\label{fig:expressions}
\end{figure}

 \begin{figure}[t]
\begin{center}
\includegraphics[width=\linewidth]{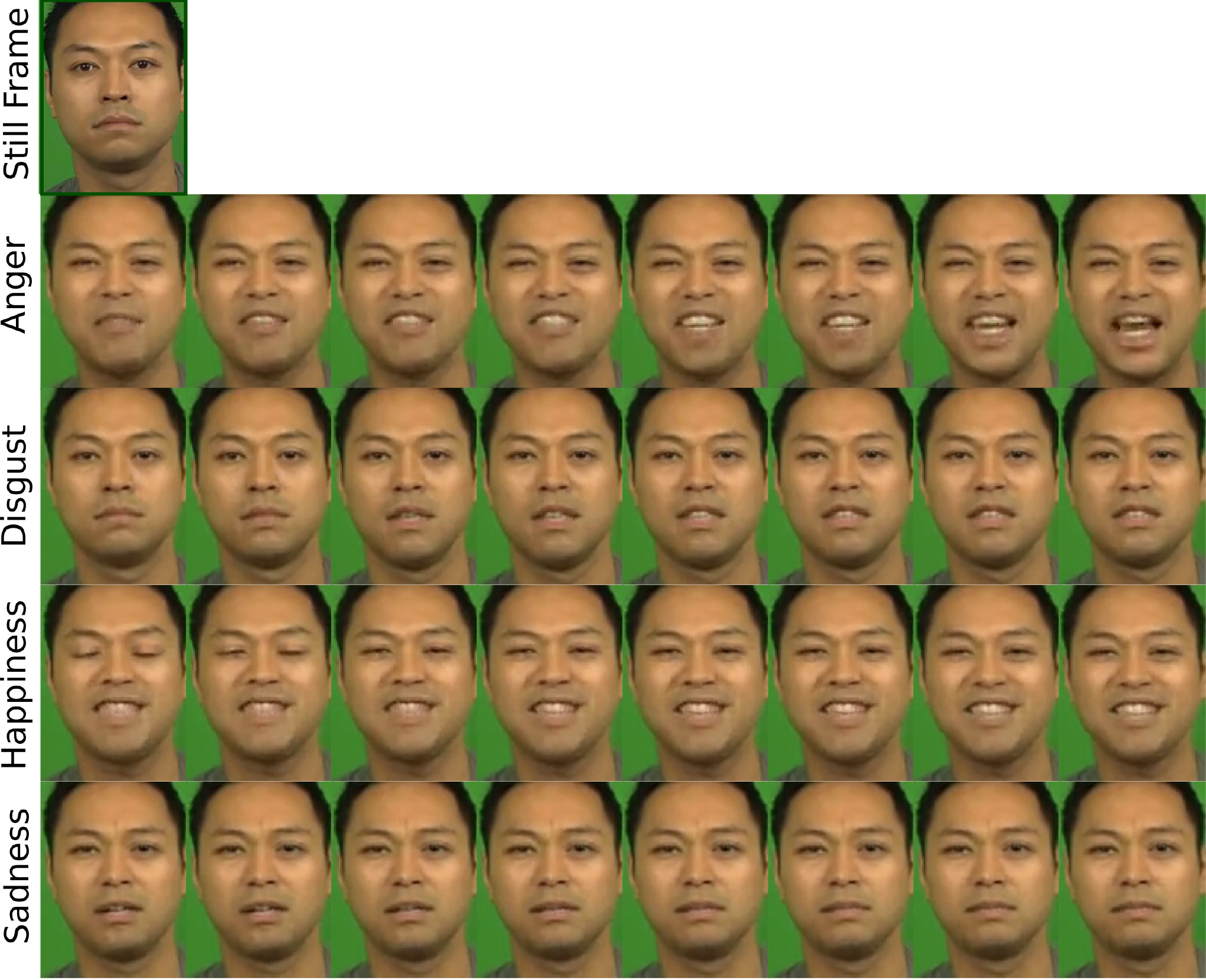}\\
\end{center}
\caption{Videos produced by the proposed method using the same image taken from the CREMA-D test set and driven by the sentence ``its eleven o'clock'' spoken with a female voice with multiple emotions.}
\label{fig:emotions}
\end{figure}

 \begin{figure*}[t]
\begin{center}
\includegraphics[width=0.76\linewidth]{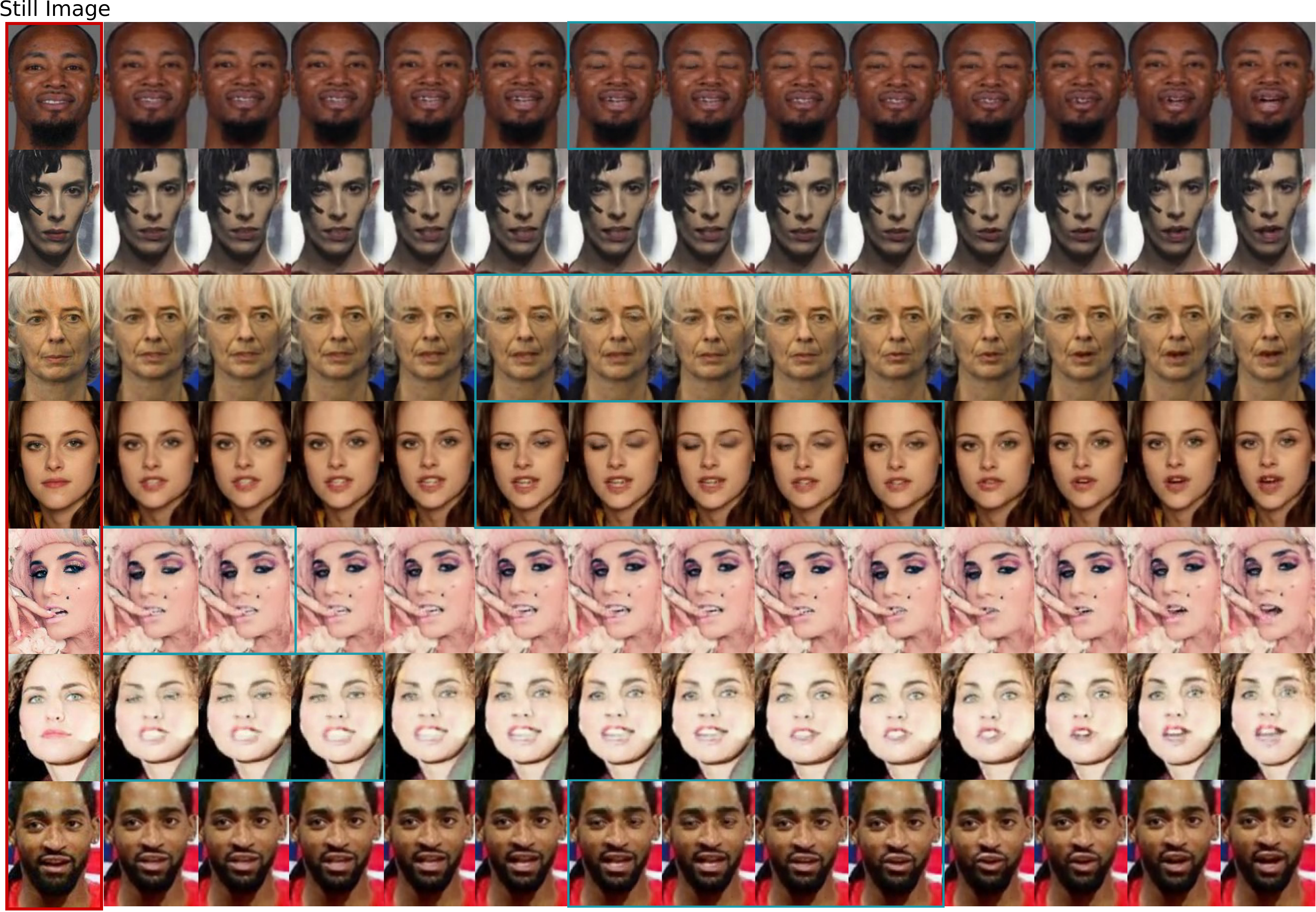}\\
\end{center}
\vspace{-11pt}
\caption{Videos produced using model trained on LRW for unseen faces taken from the CelebA dataset. The speech clip is taken from the test set of the LRW dataset and corresponds to the word "stand". Frames which contain blinking eyes are highlighted.}
\label{fig:celebA_faces}
\end{figure*}

\subsection{Qualitative Results}
\label{sec:qualitative}
Our method is capable of producing realistic videos of previously unseen faces and audio clips taken from the test set. The same audio used on different identities is shown in \figref{fig:diffface}. From visual inspection it is evident that the lips are consistently moving similarly to the ground truth video. 

Our method not only produces accurate lip movements but also natural videos that display characteristic human expressions such as frowns, blinks and angry expressions, examples of which are shown in \figref{fig:expressions}. In these examples we highlight the regions of the frames that exhibit the most movement using motion maps. These maps are obtained by calculating the optical flow between consecutive frames, reflecting the angle of movement in the hue and assigning the magnitude of the motion to the value component in the Hue Saturation Value (HSV) color-space.

The amount and variety of expressions generated is dependent on the amount of expressions present in the dataset used for training and hence faces generated by models trained on expressive datasets such as CREMA-D will exhibit a wider range of expressions. This is illustrated in \figref{fig:emotions}, where the facial expressions reflect the emotion of the speaker.

The works that are closest to ours are those proposed in \cite{Suwajanakorn2017} and \cite{Chung2017}. The former method is subject dependent and requires a large amount of data for a specific person to generate videos. There is no publicly available implementation for the \textit{Speech2Vid} method proposed in \cite{Chung2017} but a pre-trained model is provided, which we can use for comparison. For completeness we also compare against a GAN-based method that uses a combination of an $L_1$ loss and an adversarial loss on individual frames. We consider this approach as the baseline GAN-based approach. Finally, we also compare with the \emph{ATVGNet} model proposed in \cite{Chen_2019_CVPR}, which is pretrained on the LRW dataset.

\begin{figure}[t]
    \centering
    \includegraphics[width=0.77\columnwidth]{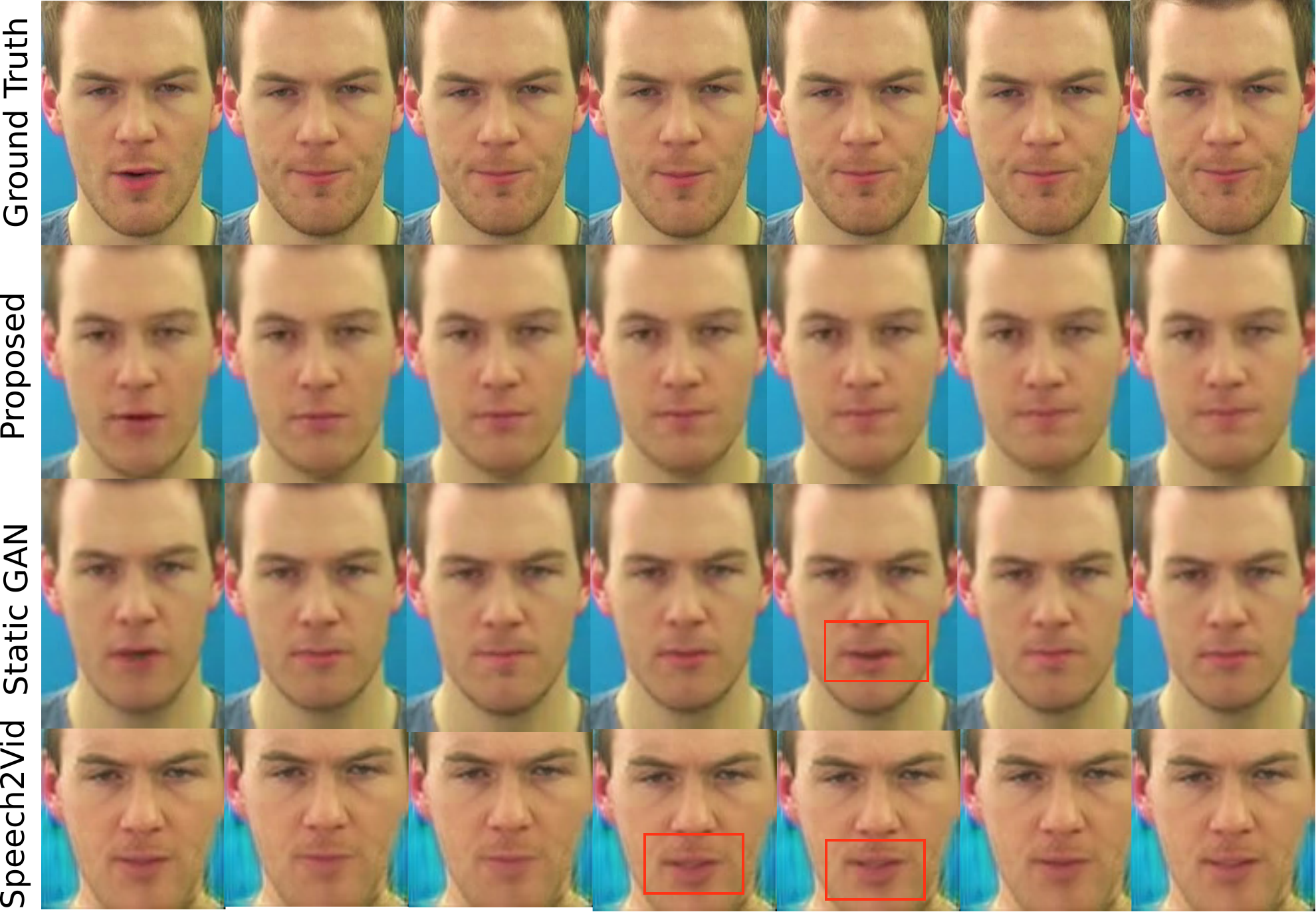}
    \caption{Example of consecutive frames showcasing the failure of static methods to produce a coherent motion. During silent periods static approaches exhibit jittery motion in the mouth.}
    \label{fig:jitterface}
\end{figure}

Since the baseline and the \textit{Speech2Vid} model are static methods they produce less coherent sequences, characterized by jitter, which becomes worse in cases where the audio is silent (e.g. pauses between words). This is likely due to the fact that there are multiple mouth shapes that correspond to silence and since the model has no knowledge of its past state it generates them at random. \figref{fig:jitterface} highlights such failures of static models and compares it to our method.

\begin{figure}[t]
 \centering
  \begin{subfigure}[b]{0.95\linewidth}
    \centering\includegraphics[width=0.99\textwidth]{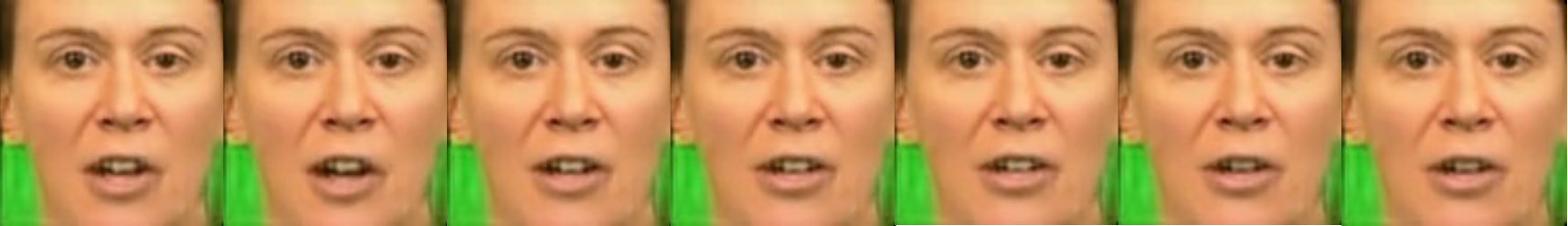}
    \caption{\label{fig:sp2vid_crema} \textit{Speech2Vid}}
  \end{subfigure}
  \begin{subfigure}[b]{0.95\linewidth}
    \centering\includegraphics[width=0.99\textwidth]{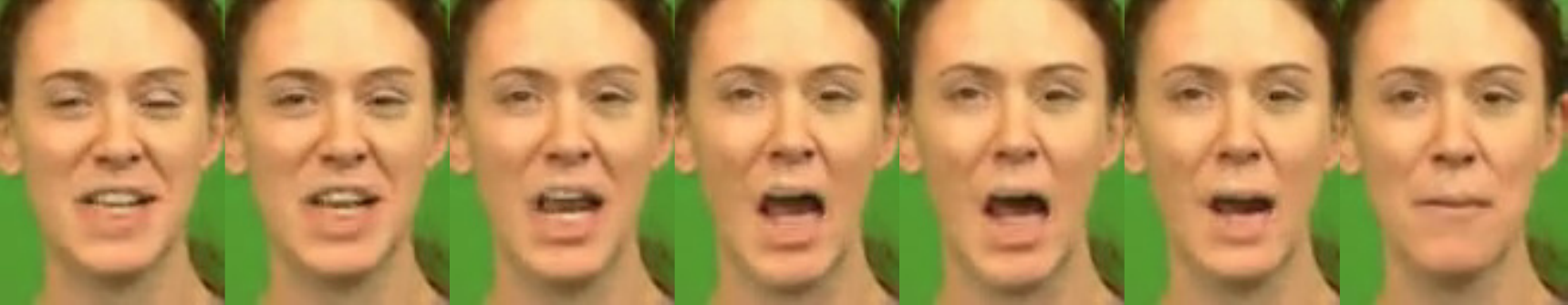}
    \caption{\label{fig:ours_crema} Proposed Model}
  \end{subfigure}
  \caption{Comparison of the proposed model with \emph{Speech2Vid}. It is obvious that \emph{Speech2Vid} can only generate mouth movements and cannot generate any facial expression.}
\label{fig:speech2vid_compare}
\end{figure}

\begin{figure}[t!]
    \centering
    \includegraphics[width=\columnwidth]{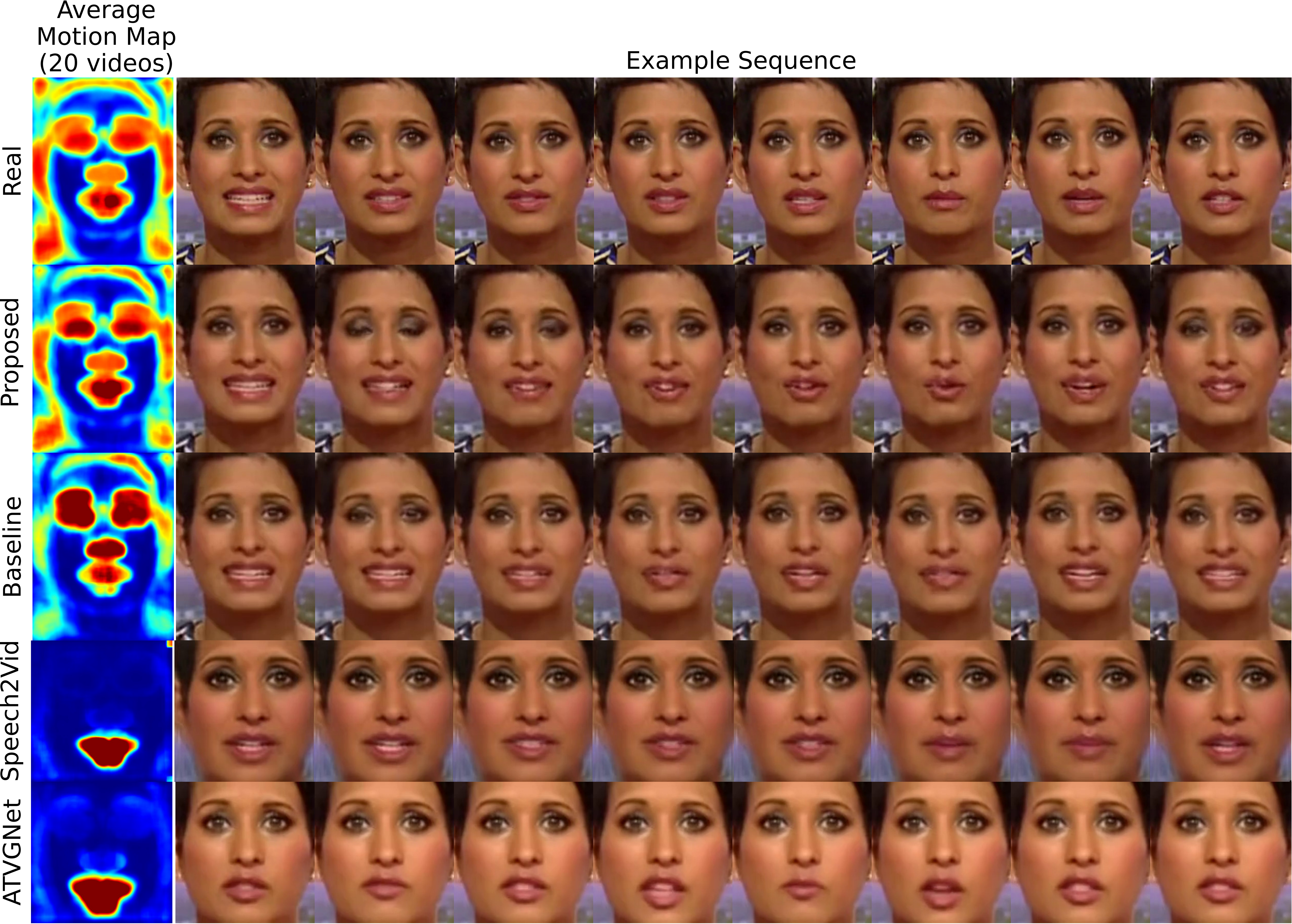}
    \caption{Average motion heatmaps showing which areas of the face exhibit the most movement. The heatmaps are an average of the magnitude of the optical flow taken for 20 videos of the same subject of the LRW dataset. An example sequence is shown next to the heatmap of each model.}
    \label{fig:average_heatmaps}
\end{figure}

The \textit{Speech2Vid} model only uses an $L_1$ reconstruction loss during training and therefore it will discourage spontaneous expressions which mostly occur on the upper part of the face. In order to examine the movement we use optical flow and create a heatmap for the average magnitude of movement over a set of 20 videos of the same subject from the LRW test set. The heatmaps shown in \figref{fig:average_heatmaps} reveal the areas of the face that are most often animated. Videos generated using our approach have heatmaps that more closely resemble those of real videos. The static baseline is characterized by considerably more motion on the face which likely corresponds to jitter. The \textit{Speech2Vid} and \emph{ATVGNet} models do not animate the upper part of the face. This means that that these methods do not capture speaker's tone and cannot therefore generate matching facial expressions. An example of this shortcoming is shown in \figref{fig:speech2vid_compare} where we compare a video generated from the CREMA-D dataset using the \textit{Speech2Vid} model and our proposed method.

\begin{table*}[t]
\small
 \centering
 \begin{tabular}{|c|l|c|c|c|c|c|c|c|c|c|c|}
 \hline
 & \multicolumn{1}{c|}{Method}& \multicolumn{1}{c|}{PSNR} & \multicolumn{1}{c|}{SSIM} & \multicolumn{1}{c|}{CPBD} & \multicolumn{1}{c|}{ACD}& \multicolumn{1}{c|}{WER}& \multicolumn{1}{c|}{AV Off.} & \multicolumn{1}{c|}{AV Conf.}& \multicolumn{1}{c|}{blinks/sec}& \multicolumn{1}{c|}{blink dur. (sec)}\\
 \hline\hline
 \small \parbox[t]{2mm}{\multirow{3}{*}{\centering \rotatebox[origin=c]{90}{GRID}}} 
 & Proposed Model & \bt{27.100} & \bt{0.818}& \bt{0.268} & 1.47 $\cdot 10^{-4}$ & \bt{23.1\%} & \bt{1} & \bt{7.4} & 0.45 & 0.36\\
 & Baseline & 27.023 & 0.811 & 0.249 & $\bt{1.42} \cdot 10^{-4}$ & 36.4\% & 2 & 6.5 & 0.04 & 0.29\\
 & \emph{Speech2Vid}&  22.662 & 0.720 & 0.255 & $ 1.48 \cdot 10^{-4}$ & 58.2\% & \bt{1} & 5.3 & 0.00 & 0.00\\
 \hline
 \small \parbox[t]{2mm}{\multirow{3}{*}{\rotatebox[origin=c]{90}{TCD}}}
 & Proposed Model & \bt{24.243} & \bt{0.730}& \bt{0.308} & \bt{1.76} $\cdot 10^{-4}$ & N/A & \bt{1} & \bt{5.5} & 0.19 & 0.33\\
 & Baseline &24.187  & 0.711 & 0.231 & $1.77 \cdot 10^{-4}$ & N/A & 8 & 1.4 & 0.08 & 0.13\\
 & \emph{Speech2Vid}& 20.305  & 0.658 & 0.211 & $ 1.81 \cdot 10^{-4}$ & N/A & \bt{1} & 4.6 & 0.00 & 0.00\\
 \hline
 \small \parbox[t]{2mm}{\multirow{3}{*}{\rotatebox[origin=c]{90}{CREMA}}}
& Proposed Model & \bt{23.565} & \bt{0.700} & 0.216 & \bt{1.40} $\cdot 10^{-4}$ & N/A & 2 & \bt{5.5} & 0.25 & 0.26\\
& Baseline       & 22.933 & 0.685 & 0.212 & $ 1.65 \cdot 10^{-4}$ & N/A & 2 & 5.2 & 0.11& 0.13\\
& \emph{Speech2Vid} & 22.190 & 0.700 & \bt{0.217} & $ 1.73 \cdot 10^{-4}$ & N/A & \bt{1} & 4.7 & 0.00 & 0.00\\
 \hline
 \small \parbox[t]{2mm}{\multirow{4}{*}{\rotatebox[origin=c]{90}{LRW}}}
& Proposed Model &\bt{23.077}  & \bt{0.757} & \bt{0.260} & 1.53 $\cdot 10^{-4}$ & N/A & \bt{1} & \bt{7.4} &  0.52& 0.28\\
& Baseline     & 22.884 & 0.746 & 0.218 & \bt{1.02} $\cdot 10^{-4}$  & N/A & 2 & 6.0 & 0.42& 0.13\\
& \emph{Speech2Vid} & 22.302 & 0.709 & 0.199 & $2.61 \cdot 10^{-4}$ & N/A & 2 & 6.2 & 0.00 & 0.00\\
& \emph{ATVGNet} & 20.107 & 0.743 & 0.189 & $ 2.14 \cdot 10^{-4}$ & N/A & 2 & 7.0 & 0.00& 0.00 \\
 \hline
 \end{tabular}
 \caption{Performance comparison of the proposed method against the static baseline and \emph{Speech2Vid} \cite{Chung2017}. A pretrained LipNet model is only available for the GRID dataset so the WER metric is omitted on other datasets. The LRW datasets contains only one word so calculating WER is not possible}
\label{tab:quantitative}
 \end{table*}
 
\subsection{Quantitative Results}
\label{sec:quantitative}
We measure the performance of our model on the GRID, TCD TIMIT, CREMA-D and LRW datasets using the metrics proposed in section \ref{sec:metrics} and compare it to the baseline and the \emph{Speech2Vid} model. For the LRW dataset we also compare with the \emph{ATVGNet} GAN-based method proposed in \cite{Chen_2019_CVPR}, for which we use the provided pretrained model. The preprocessing procedure for \emph{ATVGNet} is only provided for the LRW dataset hence we do not compare with this model on other datasets.

The results in \tabref{tab:quantitative} show that our method outperforms other approaches in both frame quality and content accuracy. For the LRW dataset our model is better not only from the static approaches but also from \emph{ATVGNet}. Our model performs similarly or better than static methods when in terms of frame-based measures (PSNR, SSIM, CBPD, ACD). However, the difference is substantial in terms of metrics that measure content such as lipreading WER. Also our method achieves a higher AV confidence, although it must be noted that based on the offset estimated using the SyncNet model our videos generated for the CREMA-D dataset exhibit a slight lag of 1 frame compared to the \textit{Speech2Vid} method. Finally, we emphasize that our model is capable of generating natural expressions, which is reflected in the amount and duration of blinks (\tabref{tab:quantitative}), closely matching those of the real videos, shown in \tabref{tab:blink_dataset}.

 We note that the \emph{Speech2Vid} and \emph{ATVGNet} methods are not capable of generating any blinks. For the \emph{Speech2Vid} model this due to using only an $L_1$ loss and for the \emph{ATVGNet} this is likely due to the attention mechanism which focuses only on the mouth since it is the region that correlates with speech. The static baseline is capable of producing frames with closed eyes but these exhibit no continuity and are characterised by very short duration as shown in \tabref{tab:quantitative}.

We further note the differences in the performance of our method for different datasets. In particular we note that the reconstruction metrics are better for the GRID dataset. In this dataset subjects are recorded under controlled conditions and faces are not characterised by much movement. Synthesized faces will mimic the motion that is present in the training videos, generating emotions and head movements. However since these movements cause deviation from the ground truth videos and therefore will be penalized by reference metrics such as PSNR and SSIM. Performance based on reconstuction metrics becomes worse as datasets become less controlled and exhibit more expressions. Another noteworthy phenomenon is the drop in audio-visual correlation, indicated by the lower AV confidence for the TCD TIMIT and CREMA-D datasets compared to GRID and LRW. We attribute to this drop in performance to the fact that the TCD TIMIT and CREMA-D are smaller datasets. It is therefore likely that the datasets do not have the sufficient data for the models to capture the articulation as well as for larger datasets.

\subsection{User Study}
\begin{figure}[t]
    \centering
    \includegraphics[width=0.85\columnwidth]{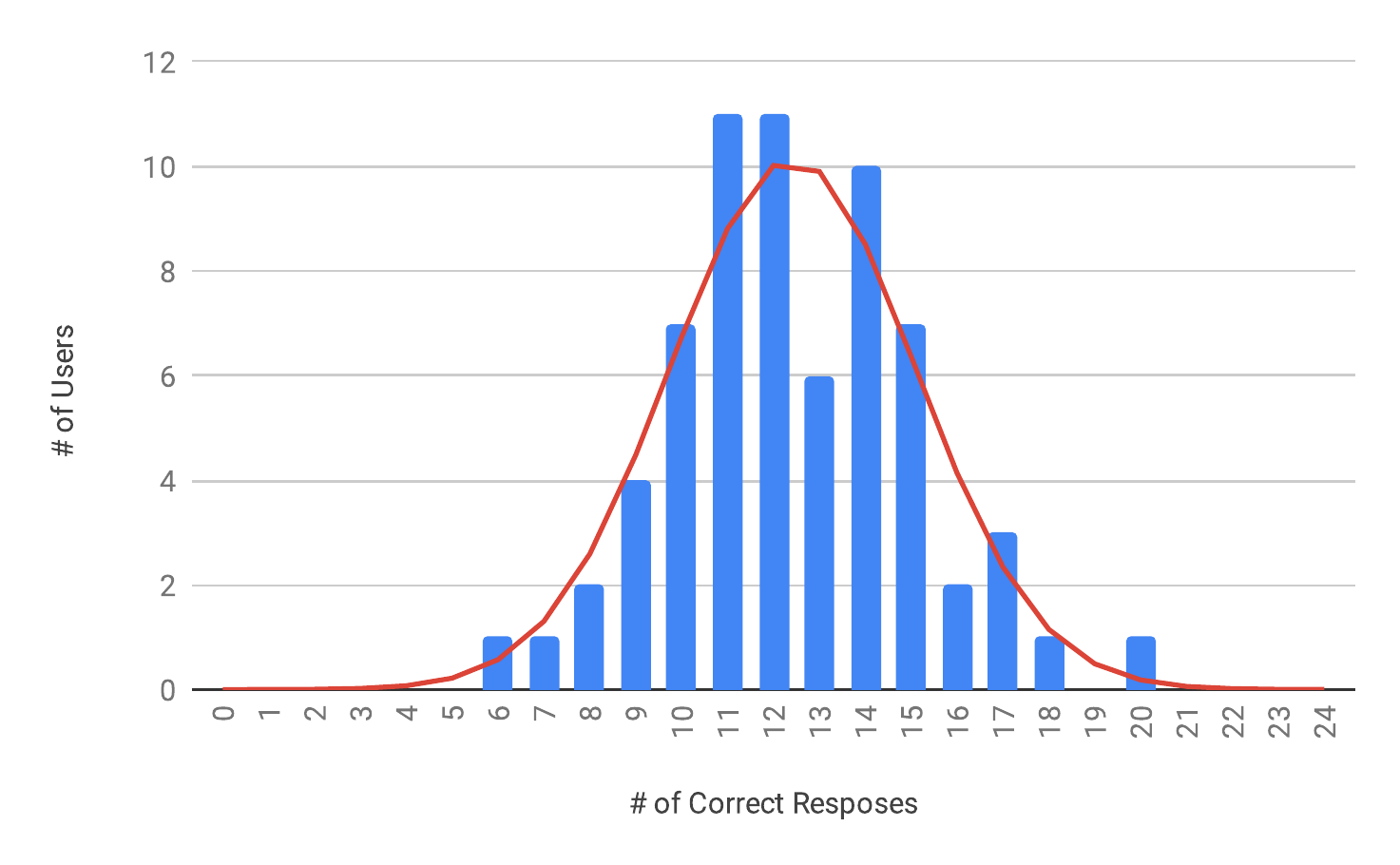}
    \caption{Distribution of correct responses of users in the online Turing test. The red line symbolizes the a Gaussian distribution with the same mean and std. dev. as the data.}
    \label{fig:turing_test}
\end{figure}

Human perception of synthesized videos is hard to quantify using objective measures. Therefore, we further evaluate the realism of the generated videos through an online Turing test \footnote{Test available \url{https://forms.gle/XDcZm8q5zbWmH7bD9}}. In this test users are shown 24 videos (12 real - 12 synthesized), which were chosen at random from the GRID, TIMIT and CREMA datasets. We have not included videos from the LRW since uploading them publicly is not permitted. Users are asked to label each video as real or fake. Responses from 66 users were collected with the average user labeling correctly $52\%$ of the videos. The distribution of user scores is shown in \figref{fig:turing_test}. 

\section{Conclusion and Future Work}
\label{sec:conclusion}
In this work we have presented an end-to-end model using temporal GANs for speech-driven facial animation. Our model produces highly detailed frames scoring high in terms of PSNR, SSIM and in terms of the sharpness on multiple datasets. According to our ablation study this can be mainly attributed to the use of a \emph{Frame Discriminator}.

Furthermore, our method produces more coherent sequences and more accurate mouth movements compared to the GAN-based static baseline and the \emph{Speech2Vid} method. This is demonstrated by a resounding difference in the WER. We believe that these improvements are not only a result of using a temporal generator but also due to the use of the \emph{Synchronization Discriminator}.

Unlike the \emph{Speech2Vid} and \emph{ATVGNet} that prohibit the generation of facial expressions, the adversarial loss on the entire sequence encourages spontaneous facial gestures. This has been demonstrated with examples of blinks, head and brow movements. Furthermore, our model is capable of capturing the emotion of the speaker and reflecting it in the generated face.

This model has shown promising results in generating lifelike videos, which produce facial expressions that reflect the speakers tone. The inability of users to distinguish the synthesized videos from the real ones in the Turing test verifies that the videos produced look natural. The current limitation of our method is that it only works for well-aligned frontal faces. Therefore, the natural progression of this work will be to produce videos that simulate in the wild conditions. Finally, future work should also focus on extending the network architecture to produce high definition video.
\section*{Acknowledgements}
We would like to thank Berk Tinaz for his help with the detection of blinks and the estimation blink duration. We also gratefully acknowledge the support of NVIDIA Corporation with the donation of the Titan V GPU used for this research and Amazon Web Services for providing the computational resources for our experiments.

\bibliography{arxiv.bib}   % name your BibTeX data base
\end{document}